\documentclass[10pt,journal,compsoc]{IEEEtran}

\ifCLASSOPTIONcompsoc
  \usepackage[nocompress]{cite}
\else
  \usepackage{cite}
\fi
\ifCLASSINFOpdf
\else
\fi

\usepackage{times}
\usepackage{epsfig}
\usepackage{graphicx}
\usepackage{amsmath}
\usepackage{amssymb}
\usepackage{physics}

\usepackage{verbatim}

\newcommand{\lspan}{{\sl span}}

\begin{document}
%
\title{The Face of Affective Disorders}
%
%
%
%

\author{Christian~S.~Pilz,~\IEEEmembership{Member,~IEEE,}
        Benjamin Clemens,
        Inka C. Hiß,
        Christoph Weiss,\newline
        Ulrich~Canzler,
        Jarek Krajewski,
        Ute~Habel,
        and Steffen~Leonhardt,~\IEEEmembership{Senior Member,~IEEE}
\IEEEcompsocitemizethanks{\IEEEcompsocthanksitem Christian S. Pilz and Ulrich Canzler are with the Research Department of CanControls GmbH, Aachen, Germany. Christian S. Pilz is also visiting scholar with the Department of Psychiatry, Psychotherapy and Psychosomatics, University Hospital, Aachen, Germany.\protect\\
E-mail: pilz@cancontrols.com or cpilz@ukaachen.de

\IEEEcompsocthanksitem Ute Habel, Benjamin Clemens and Inka C. Hiß are with the Department of Psychiatry, Psychotherapy and
Psychosomatics, University Hospital, Aachen, Germany.
\IEEEcompsocthanksitem Jarek Krajewski is with the Department of Economy and Psychology, University of Applied Science, Cologne, Germany.
\IEEEcompsocthanksitem Steffen Leonhardt and Christoph Weiss are with the Chair of Medical Information Technology (MedIT), RWTH Aachen University, Aachen, Germany.}
\thanks{Manuscript received April 19, 2005; revised August 26, 2015.}}

%
%

\markboth{Journal of \LaTeX\ Class Files,~Vol.~14, No.~8, August~2015}%
{Shell \MakeLowercase{\textit{et al.}}: Bare Advanced Demo of IEEEtran.cls for IEEE Computer Society Journals}
%



\IEEEtitleabstractindextext{%
\begin{abstract}
We study the statistical properties of facial behaviour altered by the regulation of brain arousal in the clinical domain of psychiatry. The underlying mechanism is linked to the empirical interpretation of the vigilance continuum as behavioral surrogate measurement for certain states of mind. Referring to the classical scalp-based obtrusive measurements, we name the presented method \textit{Opto-Electronic Encephalography (OEG)} which solely relies on modern camera-based real-time signal processing and computer vision. Based upon a stochastic representation as coherence of the face dynamics, reflecting the hemifacial asymmetry in emotion expressions, we demonstrate an almost flawless distinction between patients and healthy controls as well as between the mental disorders depression and schizophrenia and the symptom severity. In contrast to the standard diagnostic process, which is time-consuming, subjective and does not incorporate neurobiological data such as real-time face dynamics, the objective stochastic modeling of the affective responsiveness only requires a few minutes of video-based facial recordings. We also highlight the potential of the methodology as a causal inference model in transdiagnostic analysis to predict the outcome of pharmacological treatment. All results are obtained on a clinical longitudinal data collection with an amount of 99 patients and 43 controls.
\end{abstract}


}

\maketitle

\IEEEdisplaynontitleabstractindextext

%
\IEEEpeerreviewmaketitle

\ifCLASSOPTIONcompsoc
\IEEEraisesectionheading{\section{Introduction}\label{sec:introduction}}
\else
\section{Introduction}
\label{sec:introduction}
\fi

\begin{figure*}[!h]
\centering
  \includegraphics[width=1.\textwidth]{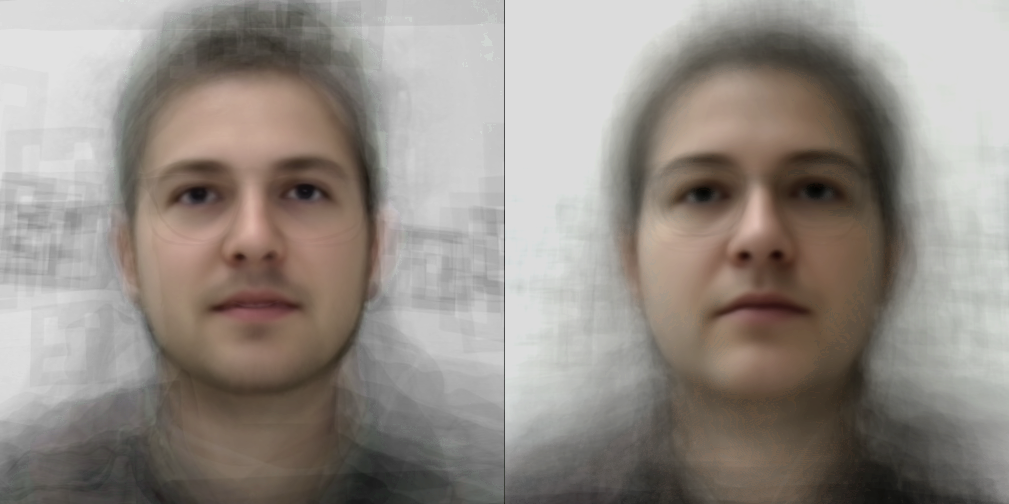}
  \caption{The neutral mean faces of the healthy control group and the patients are visualized. The left face corresponds to the control group and the right face to the patient group. The mean face of individuals diagnosed with affective disorders were perceived as having more hanging mouth corners, raised eye browns, more swollen pale faces and a dull gaze. Affective disorders also were associated with female gender, sad emotion, less trustworthy, less leadership quality and less attractive face appearance. As a whole, this was also associated with illness. The detailed results of the corresponding questionnaire are described in subsection \ref{subsec:cues}}
   \label{fig:the_face_of_affective_disorders}
\end{figure*}

\IEEEPARstart{M}{}ental disorders are among the top ten leading causes of public health burden, worldwide and there has been no evidence of a global reduction of this burden in the last 30 years \cite{Lancet2022}. The global number of disability adjusted life years (DALYS) and the proportion of global DALYS attributed to mental disorders increased. It has become clear that the classification of mental disorders by taxonomies such
as the Diagnostic and Statistical Manual  of Mental Disorders (DSM) or the International Classification of Diseases
(ICD) is not sufficient as they do not reflect relevant neurobiological and behavioural mechanisms.
Therefore, in the US the National Institute of Mental Health started the Research Domain Criteria (RDoC) project to establish
a research classification system for mental disorders that is based on
neurobiological findings and observational behaviour \cite{Cuthbert2013}.  In Europe, the Roadmap
for  Mental  Health  Research  in  Europe  (ROAMER)  project    paved  the  way  for
developing  a  comprehensive  and  integrated  mental  health  research  agenda  which  is  successfully
featured by the German Federal Ministry of Research and Education initiative e:Med (for example:
IntegraMent - Integrated Understanding of Causes and Mechanisms in Mental Disorders) \cite{Haro2014}. Today,  we  have  access  to  ground-breaking  advances  in  biological  and  brain  sciences;  biomarkers
from ‘-omics’ research, developments in brain mapping such as the connectome, fast genome-wide
association  studies,  next  generation  DNA  sequencing,  affective  computing,  eHealth,  cognitive
behavioural therapy and large scale research infrastructures \cite{Yuste2014}. Taking advantage of such
developments will produce a larger body of evidence along the entire translational pipeline starting from
biological mechanisms to behavioural systems, resulting in new clinical approaches and preventative interventions,
leading to more sustainable and individualized treatments \cite{Wykes2015}. However, for the moment
patient-tailored treatment has not yet arrived in the clinical setting.  The analysis of genetic factors and
their interaction with environmental factors has delivered valuable sets of predictors. Utilizing genetic profiles, together with the environmental factors including lifestyle, it is possible to
develop predictors classifying individuals at risk; this way, establishing preventive
strategies to reduce the risk of developing a mental disorder \cite{Topol2014}. 
Classical statistical concepts like  null-hypothesis  testing  have  struggled in  dealing  with  objectively  measurable  endophenotypes  derived  from  huge  datasets  and in extrapolating  patterns  from  one  set  of  data  to  another. While  these  tests  often  determine
differences between affected versus healthy subjects or treatment versus placebo group, it turned out that they do not help in finding a differential diagnosis or the right treatment among numerous competing treatment groups   \cite{Bzdok2018}.  In  contrast,  machine  learning  uncovers  general concepts  underlying  a  series  of  observations  without  explicit  instructions to reveal biological subgroups in patients \cite{Hahn2017,Bzdok2018}.
In mental disorders with complex possible treatment combinations, where differential diagnosis and disease trajectories may fall into more than two categories, multiclass learning approaches must be used. Unique combinations of behavioral, genetic, neural, hormonal and environmental
characteristics play a role in the pathogenesis of mental illness. Fusing all these types of
data and levels of analysis to produce statistically robust large-scale
models is an important challenge for the future. Machine learning  has  been  successfully  applied  in  various  research areas in recent years, such as
language  processing,  speech  recognition  or  computer  vision and is suitable to offer a framework of mechanisms
to enable clinical prediction on an individual level \cite{Shamout2021}. It is expected that the combination of genetic data with physiological
biomarkers, behaviour and clinical factors with the concept of learning algorithms may enable the identification of disease-specific  biological  aspects  that  help  to  allocate  patient  subgroups  to  specific  treatment
options, and to identify subjects at risk. In this way, using artificial intelligence could provide specific
treatment  options  or  preventive  strategies  tailored  to  the  individual  person,  across  the  common
mental disease classifications.

\begin{figure*}[!h]
\centering
  \includegraphics[width=0.85\textwidth]{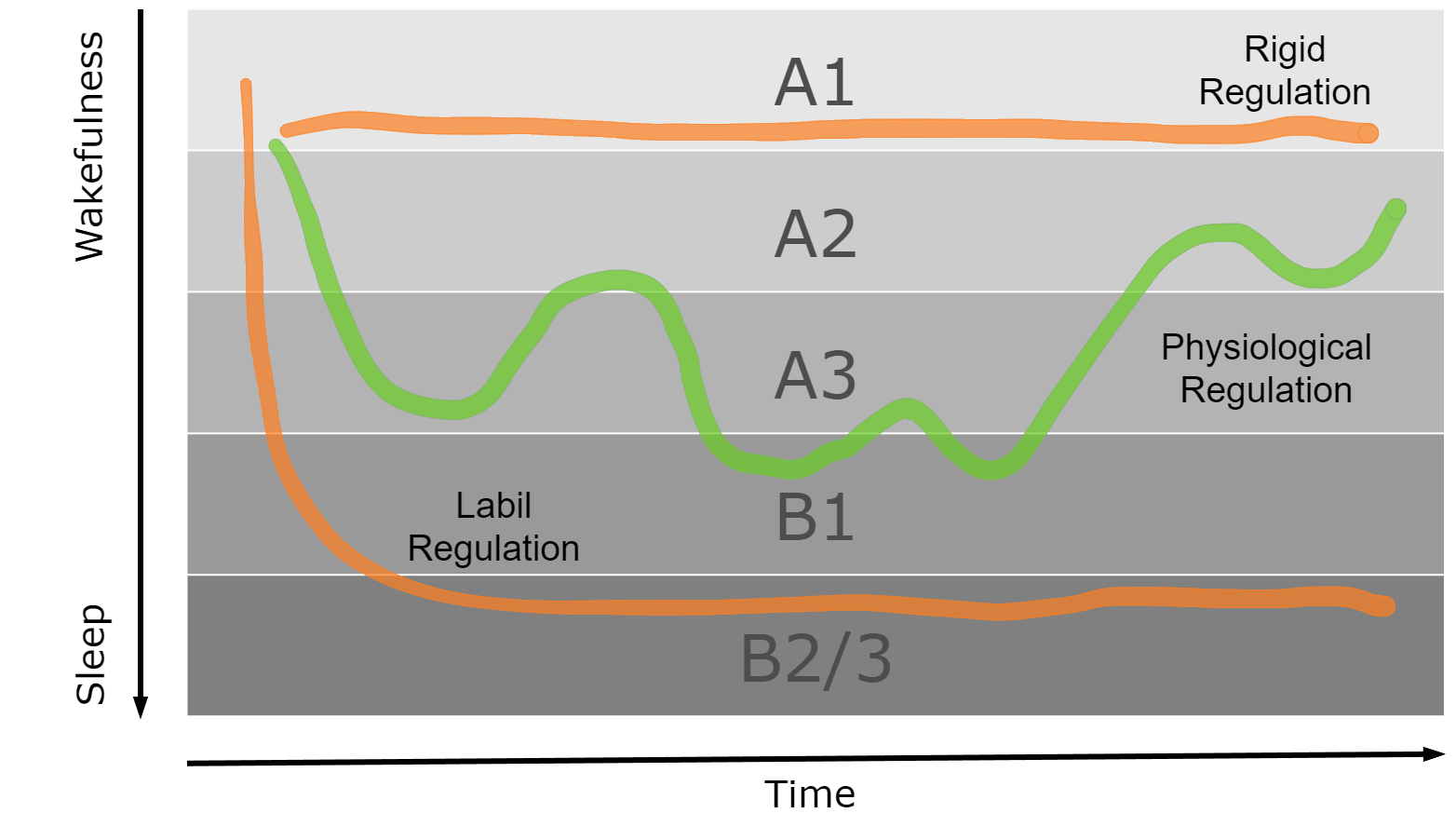}
  \caption{Brain arousal regulation of the vigilance continuum as synopsis of defined classifications of the EEG-stages occurring between relaxed wakefulness during closed eyes and deep sleep \cite{Ulrich2013}. The stages A1 to B2 have first been introduced by Bente and Roth \cite{Bente1964}. A decrease in vigilance can lead to two different behaviors, 1) the organism decides to go to sleep and the vigilance reverts to sleep stages or, 2) the organism exhibits auto-stabilization behavior to counter regulate their vigilance level. If this natural physiological regulation is permanently disturbed, pathological symptoms can develop. The rigid regulation is often observed during depressive episodes and the labile regulation during manic episodes \cite{Hegerl2014}. A statistical link to facial dynamics was studied recently \cite{Pilz2020}.}
   \label{fig:vigilance_regulation}
\end{figure*}

\section{Contribution}
\label{sec:Contribution}

In this context, this study pays particular attention to the behavioural manifestations of affective
disorders, and primarily in what is probably the most pronounced modality in everyday life, especially
with respect to unconscious behaviour, that of facial expression. Going beyond machine diagnosis of
depression in affective computing, which has been developed in previous studies \cite{Kacem2018, Harati2020},
we show that the measurable affective state estimated by means of computer vision contains far more information
than the pure categorical classification.  It is a further objective to emphasize that, in the overall
context of mental illness, the patient's emotional state is an essential key factor to understanding the
underlying disease. Although this appears to be known in the clinical field, we believe that a broader and
more public awareness of this issue and its background can lead to constructive problem-solving approaches.  In a general and
interdisciplinary sense, this work with its theoretical background and technical realisation attempts to
contribute in trying to understand affective syndromes by experiential measures and multimodal affectives responses, in the spirit of \textit{the rise of
affectivism} \cite{Dukes2021}.
First, we review the connection between facial behaviour and the psychiatric field.  Next, we show that the facial expression manifestation of affective
disorders can be attributed to classical brain arousal regulation (a phenomenon formerly known as vigilance).
Thereafter, we explain how behavioural face dynamics can be formulated mathematically.  Here, we expand the
terminology of shape geodesics to further take into account the temporal dimension, which until now has not
received much consideration. Thus, we look not only at what the person is doing at a given moment, but
also how they do it and what the connections between the individual components of the observable
process are \cite{Granger1969}. This assumption is based on the theory of hemifacial asymmetry in emotion expressions \cite{Mandal1998}. Quoting Aristotle, \textit{"The whole is more than the sum of its parts"}. 
To evaluate the algorithms empirically\footnote{We published the code for reproducing all results on the corresponding author\textquotesingle s public repository.}, we present a new clinical data collection\footnote{The data collection will
be made available upon request in compliance with the underlying data protection policies to qualified individuals our groups to support broader research and enable
collaboration in this field.}. The visual differences between the patients and the healthy controls are examined and presented using the collected video data. We also show how they are perceived by other people (see Fig. \ref{fig:the_face_of_affective_disorders} for the corresponding mean faces). Based on the clinical diagnosis the classic categorical
classification will be presented. This is followed by the computer vision-based measurement of the human face,
whose construction we justify and compare to known methods.  In contrast to the static observation of facial
expressions, the dynamic behaviour can also distinguish between the disease categories of depression and
schizophrenia and not only between healthy controls and patients.  In further experiments, the connection between facial
dynamics and symptom severity will be examined.  In the final set of experiments, a model of causal inference
is evaluated to obtain patient-tailored drug treatment recommendations.

\section{Background}
\subsection{The Facial Expressions}
\label{sec:soul}
Attempts to define the concept of feelings or emotions go back at least to Aristotle, who described them as
situation-specific perceptual interpretations \cite{Aristoteles1913}.  Darwin also dealt with the expression
of emotions in humans and animals. He found that certain expressive movements could be explained through
protective functions or deterrence \cite{Darwin1872}. Later theories emphasized the role of central nervous processes and regarded emotional excitement or arousal as a function and interaction of cortical and subcortical processes. More recent theories also assume cognitive states that are independent of physiological arousal as a sufficient criterion for the expression of emotions \cite{Werner1982}.  In the seminal work \textit{The biology of human behaviour}, Eibl-Eibesfeldt claimed that the expressive movements associated with emotions are the same in all cultures \cite{Eibl_Eibesfeldt1975}.  Some authors assume the existence of so-called basic emotions \cite{Tomkins1962, Izard1977, Ekman1992_2}. Among these, Ekman also posits that expression and meaning are culture-independent, with a strong reference to Darwin\textquotesingle s universality of facial expressions of emotion.  However, current views regard these findings rather critically and show that emotions tend not to be cross-cultural constants \cite{Jack2014}.  The evolution of emotions, in a genuine biological sense, can be better traced back to the \textit{fight or flight} response. Essentially, this reduces the possible set of base emotions, with more complex emotions reflecting specific cultural idiosyncrasies \cite{Cannon1929, Bracha2004}. Regardless of cultural characteristics, it is beyond doubt that one\textquotesingle s facial expression---interplay of eye, mouth and face muscles---reflects one\textquotesingle s feelings, emotions and thoughts:  it is a window to one\textquotesingle s inner state.  Leonhard has described facial expression as the \textit{language of expression of the soul} \cite{Leonhard1968}. As recognized very early on,
affects are expressed in the motor system, consequently also in facial expressions, gestures, gait and posture \cite{Bleuler1911}. A change in a person\textquotesingle s emotional world or psyche is therefore also reflected in a change of facial expression. Pideret also described the role of facial expression in neuropsychiatry as well as in psychology as one of the most important sources for the detection of pathological of mental states \cite{Pideret1896}. Psychiatrists usually judge the affect in patients as more flattened, labile or inadequate. The facial expression behaviour seems partly impoverished, bizarre or uncontrolled up to the point of given stereotypes. Numerous studies exist today that have attempted to use computer vision and machine learning to investigate affective disorders via facial expressions \cite{Grabowski2019}.
Current research focuses on the dynamic interpretation of facial features and also predicting the symptom severity of the disease \cite{Song2022}.


\begin{figure*}[!h]
\begin{center}
\includegraphics[scale=.76]{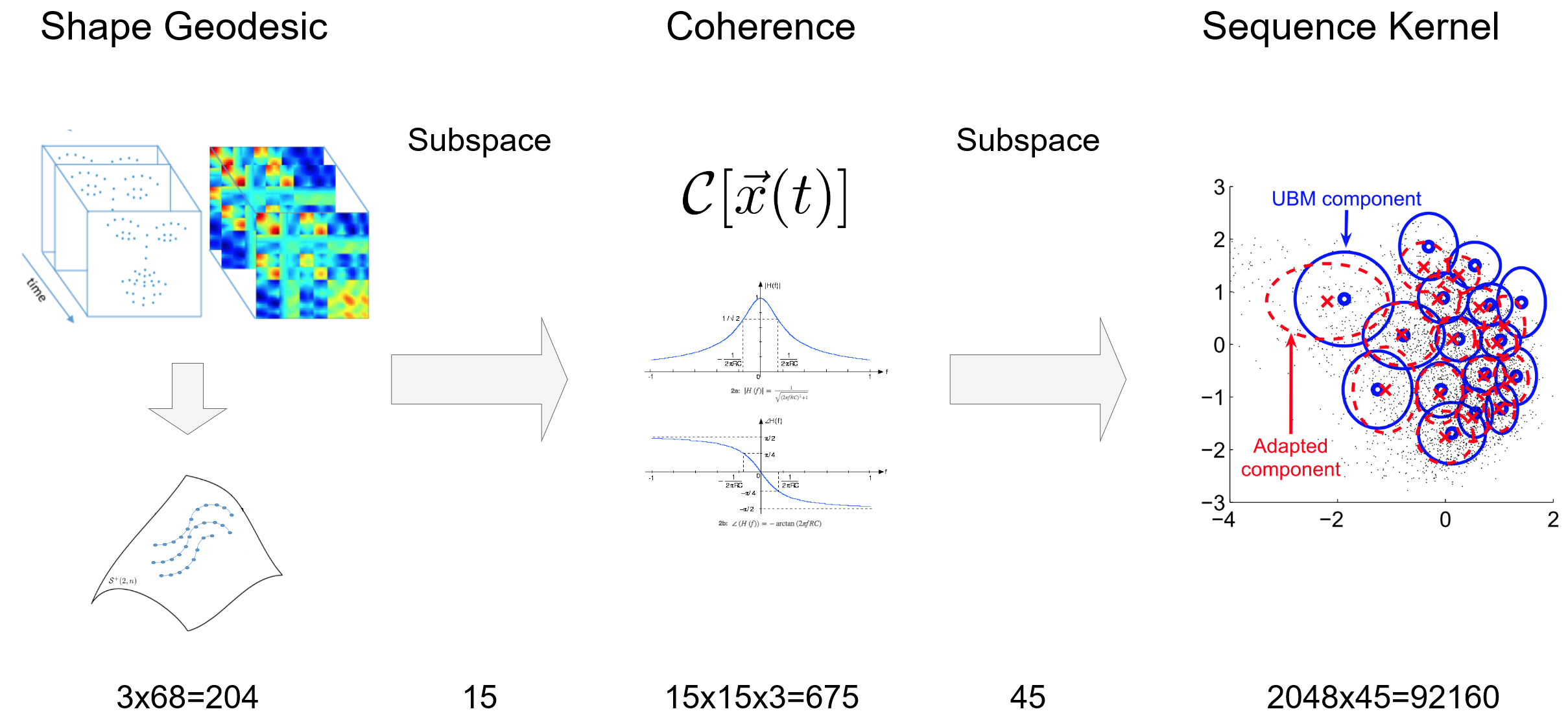}
\caption{Illustration of the computational pipeline for the shape-based coherence sequence kernel. Given a set of registered facial landmarks the trajectory of the shape geodesic will be modeled as VAR model over sliding blocks. Each block is gradually accumulated into the pre-computed background mixture model. At the end the kernel is then calculated via an adaptation step and fed to the Gaussian process regression. A subspace projection was carried out between the computation steps in order to reduce the overall complexity.}
\label{fig:shape_geodesic_pipeline}       
\end{center}
\end{figure*}

\subsection{The Vigilance}
\label{sec:vigilance}
Many biological systems can be described relatively well as open thermodynamic processes \cite{Schroedinger1944}. The alternation between determinism and complexity of this autopoietic system \cite{MaturanaVarela1973} by means of its dissipative structures \cite{Prigogine1977} on an abstract level involves merely certain different states of energy \cite{Pilz2020}. The transformation of energy into a new state is nothing more than information \cite{Shannon1948_2}. There is a mutual relationship between physical, psychological, and social aspects of health. It is evident that health is closely linked to individual and collective value systems and behavior patterns that manifest in personal lifestyles. Life is a continuum of maintaining equilibrium \cite{Hurrelmann1989}. Continuous disturbances, however, can lead to certain diseases. The key question is which information indicates an unbalanced metabolism at all and whether it is possible to use this to make objective statements about the possible course of the disease, suitable treatments and the chances of recovery.\newline
Affective disorders belong to the clinical discipline of neuropsychiatry. Attempts have been made to understand the complex nature of these diseases and their causes for a long time. In many cases, these are processes that have developed over many years and which have often become profoundly consolidated through ongoing difficult emotional and social circumstances.  Biologically, the local protein synthesis inside the brain gets thrown out of balance, which leads to a drastic change in brain arousal regulation. This physiological phenomenon is known as \textit{vigilance}, whose origin dates back to the seminal work of Head \cite{Head1923}. In its original formulation, vigilance is neither a function nor a performance, but is rather described as a continuum. It has more a semantic meaning like the \textit{explanatory principles} \cite{Bateson1979} and therefore can be regarded as a \textit{theoretical construct} \cite{Carnap1956}. Accordingly this construct is primarily conceived for scientists, who are interested in the reality behind the empirical phenomena. Vigilance is often used as a summarizing term in an interdisciplinary context for discrete measurements like sustained attention, selective attention or alertness. However, these terms reflect certain measurable performances and the mechanism of enabling a certain performance respectively. This is also represented by the early protagonists and it should be explicitly pointed out in order to avoid corresponding misinterpretations \cite{Ulrich1988, Ulrich2013}. Electroencephalography (EEG) changes of vigilance by a complex spatio-temporal process were first presented by Loomis and Davis\cite{Davis1938,Loomis1937}. In this early perspective, a distinction was initially only made between the two states of wakefulness and sleep. Some decades later, Bente and Roth \cite{Bente1964,Roth961} explained the states in a more detailed representation by dividing these into substages (see Figure \ref{fig:vigilance_regulation}). With the rapid development and increasing performance of the semiconductor industry the first machine learning approach for the classification of these stages was realized given the work of Ulrich, Olbrich and Hegerl \cite{Hegerl2014,Ulrich2013,Ulrich1986}. Basically, vigilance is usually measured as the connectivity between the visual cortex and the frontal lobe in terms of EEG alpha waves in the classical sense. Alpha waves are known to be involved in the deactivation of the corresponding brain region \cite{Ulrich2013}. However, there are also several surrogate measurements linked to this arousal dynamic since it is not always possible to incorporate obtrusive sensor like EEG
or the more popular stationary sensing devices like Magnetic resonance imaging (MRI) technology \cite{Kloesch2022}. More recent studies have analyzed vigilance regulation while one\textquotesingle s eyes are open. Here the statistical relation between the facial behaviour, the vegetative nervous system and performance-based vigilance measurements are used as a multivariate construct to directly predict brain arousal regulation as an indicator for estimating tiredness \cite{Pilz2020}. In this paper, the term vigilance will be also used as a surrogate measure which is further investigated in the domain of observable behaviour from the human face. In essence, if there exists a labile regulation which is associated to affective disorders like depression, then the facial dynamics should be also significant different compared to the state of natural physiological regulation found in healthy people (see Figure \ref{fig:vigilance_regulation} for the regulation schematics).

\section{The Model Space}
\label{sec:model}
Observing the human face in interpersonal communication sending signals which one would like to interpret. Signals in the proper sense are nothing more than functions in mathematical spaces. Functions, on the other hand, can be described by the changes in their variables. For a given problem, the resulting question will be which kind of functional can describe a suitable solution. For the registration of the face shape, a solution via the representation as an inverse problem yields to a nonlinear least square problem which has proven to be powerful. In general, a shape is a point in a high-dimensional, nonlinear manifold, called a shape space sharing the properties of the Riemannian geometry. Kendall pioneered the study of shape for labeled point sets \cite{Kendall1989}. Therefore, the properties of such a representation are relatively well understood. The Riemannian structure can only be considered locally as Euclidean. The temporal consideration of the shape on the manifold is a geodesic path represented by the corresponding Lie algebra. In essence, we are seeking an analytic solution for the face shape as equivalence class describing the space of diffeomorphisms as function of time. A schematic overview of the model space which can solve this problem is illustrated in Fig. \ref{fig:shape_geodesic_pipeline} and will be described in the following two subsections.


\subsection{Shape Geodesic}

\begin{figure}[h!]
\begin{center}
\includegraphics[scale=0.95]{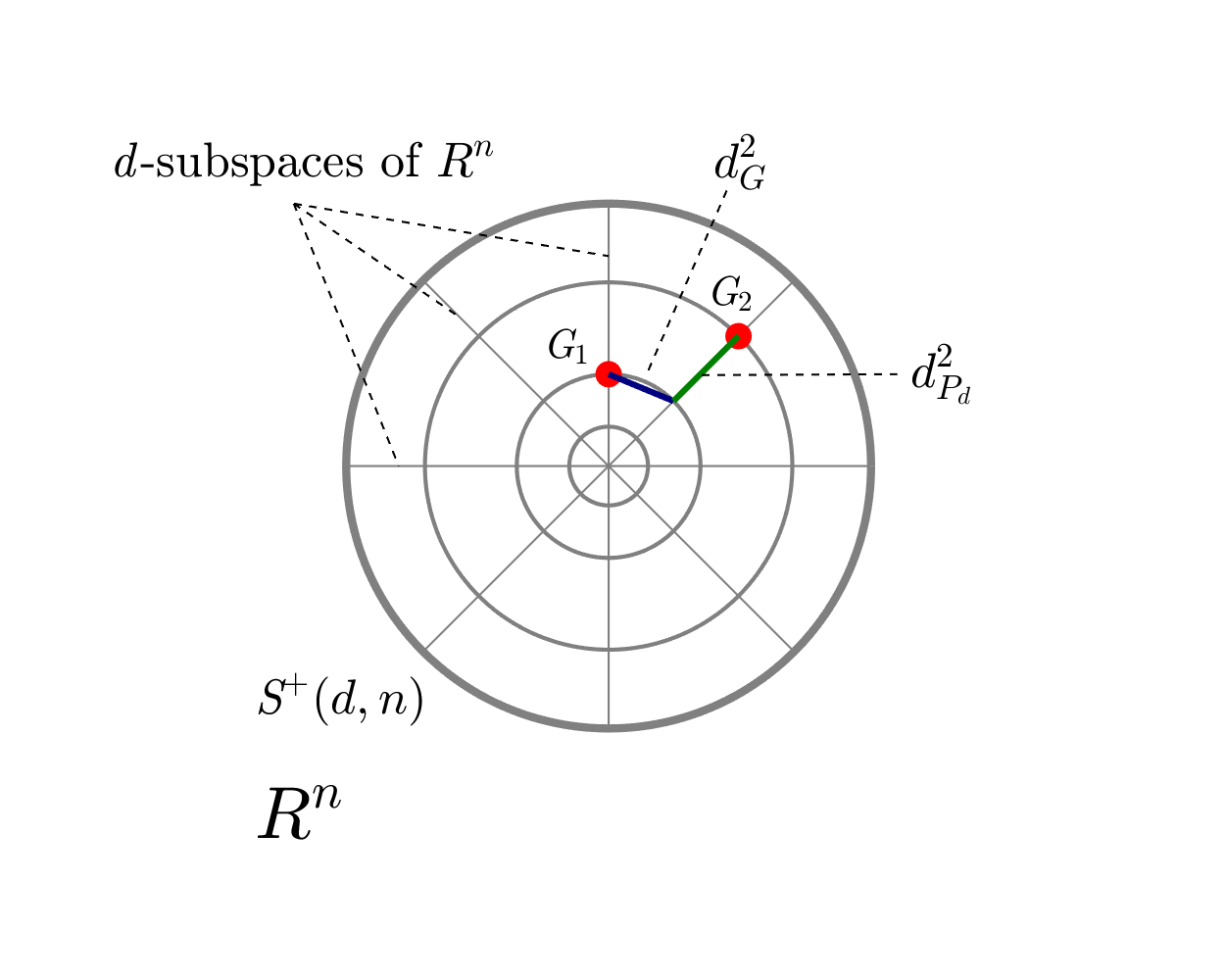}
\caption{ A graphical top-down view of the positive semi-definite
cone $\mathcal{S}^+(d,n)$. The matrices $G1$ and $G2$ are distributed in $\mathbb{R}^d$. Their approximated distance consists of the 
squared Grassmann distance $d^2_{G}$ and the squared Riemannian distance in
$d^2_{P_{2}}$.}
\label{fig:postive_cone}       
\end{center}
\end{figure}

Given a sequence of vectors representing the shape of an object with a certain fixed amount of landmarks. Arguably, the intrinsic change of the shape representation, the shape dynamic, needs to be invariant to the action of the special Euclidean group SE(3) in order to be able to execute any kind of comparison usually done with a propriety distance measurement.
Consider an arbitrary sequence of landmark configurations $\tilde{X}:=\{X_i: i=1,...,\tau\}$ representing the non-rigid motion of the object\textquotesingle s shape as a function of time, where each $X_{i}$ $(0 \leq i\leq \tau)$ is an $n \times d$ matrix of rank $d$ encoding the positions of the $n$ distinct landmark points $\vec{p} \in  \mathbb{R}^d$. Usually with $d \in  \{2,3\}$, the two- or three-dimensional Euclidean space. At each time step $i$, a common choice to incorporate certain necessary invariance properties to the shape representation would yield to the matrix of pairwise distances between the landmarks of the same shape augmented by the distances between all the landmarks and their \textit{center of mass} $p_0$. 

\begin{subequations}
\begin{align}
\mathbb{D} & = (d_{ij}); \\
d_{ij} & = d_{ij}^2 \;=\; \lVert p_i - p_j\rVert^2
\end{align}
\end{subequations}

where $\|\cdot\|$ denotes the Euclidean norm on $\mathbb{R}^d$.

\begin{equation}
\mathbb{D} = \begin{bmatrix}
0 & d_{12}^2 & d_{13}^2 & \dots & d_{1n}^2 \\
d_{21}^2 & 0 & d_{23}^2 & \dots & d_{2n}^2 \\
d_{31}^2 & d_{32}^2 & 0 & \dots & d_{3n}^2 \\
\vdots&\vdots & \vdots & \ddots&\vdots&  \\
d_{n1}^2 & d_{n2}^2 & d_{n3}^2 & \dots & 0 \\
\end{bmatrix} 
\end{equation}
The Euclidean distance matrices are related to Gram matrices \cite{Gower1982}
where $\| \cdot \|$ denotes the norm associated to the $l^2$-inner product $\langle \cdot , \cdot \rangle$. The entries of the Gram matrix $G := XX^T$ are the pairwise inner products of the points $p_1, \ldots, p_n$,
\begin{equation}
\label{eq:gram}
G=XX^T = \langle p_i, p_j\rangle, \; \  1 \leq i, j \leq n \; ,
\end{equation}

\noindent with the linear relationship given the equality
\begin{equation}
\mathcal{D}_{ij} =
\langle p_i, p_i \rangle
- 2 \langle p_i, p_j \rangle
+ \langle p_j, p_j \rangle, \; \ 0 \leq i, j \leq n \;.
\end{equation}
Gram matrices are $n \times n$ positive semidefinite matrices of rank $d$.
The Riemannian geometry of the space of such matrices is given by the positive semidefinite cone $\mathcal{S}^+(d,n)$. The Grassmann manifold $\mathcal{G}(d,n)$ of $d$-dimensional subspaces in $\mathbb{R}^n$ is invariant under transformations of the special Euclidean group SE(3) resulting in an affine invariant shape representation given by the subspace $U$ spanned by the columns of $X$ \cite{Begelfor2006}
\begin{equation}
U=\lspan(X).
\end{equation}
For two subspaces 
\begin{equation}
\mathcal{U}_1 = \lspan(U_1) 
\end{equation}
and 
\begin{equation}
\mathcal{U}_2=\lspan(U_2) \in \mathcal{G}(d,n), 
\end{equation}
the geodesic curve connecting them is given by
\begin{equation}
\label{eq:GeoGrass}
\lspan(U(t))= \lspan(U_{1}\cos(\Theta t)+M\sin(\Theta t)) \; ,
\end{equation}

\noindent where $\Theta$ is a $d \times d$ diagonal matrix formed by the principal angles between $\mathcal{U}_1$ and $\mathcal{U}_2$. The matrix $M$ is given by the formula $M=(I_n-U_{1}U_1^T)U_2 F$ and $F$ is the pseudoinverse $diag(\sin(\theta_1),\sin(\theta_2))$. The Riemannian distance between the subspaces $\mathcal{U}_1$ and $\mathcal{U}_2$ is given by
\begin{equation}
d^2_{\mathcal{G}}(\mathcal{U}_1,\mathcal{U}_2)=\|\Theta\|^2_F \;  .
\label{eq:distGrass}
\end{equation}
The polar decomposition of a $n \times d$ matrix $X$ of rank $d$, 
\begin{equation}
X = UR
\end{equation}
with 
\begin{equation}
R = (X^T X)^{1/2}
\end{equation}
yields to the Gram matrix $XX^T$ as $UR^2U^T$. The columns of the matrix $U$ are orthonormal and the matrix $R$ is postive-definite. The polar decomposition
maps from the product of the Stiefel manifold $V$ and the cone of the positive-definite. matrix $P$ to the manifold $\mathcal{S}^+(d,n)$
\begin{alignat*}{2}
\Pi : & \mathcal{V}_{d,n} \times  \mathcal{P}_d \rightarrow \mathcal{S}^+(d,n) \\
  &(U,R^2)\mapsto UR^2U^T \;.
\end{alignat*}
Connecting two Gram matrices $G_1 = U_1R_1^2U_1^T$ and $G_2 =  U_2R_2^2U_2^T$ in $\mathcal{S}^+(d,n)$ is given by the curve
\begin{equation}
\mathcal{C}_{G_1\to G_2}(t)=U(t)R^2(t)U^T(t), \forall t \in [0,1] \; ,
\label{eq:geopsd}
\end{equation}
\noindent with $R^2(t)=R_{1}\exp(t\log R_{1}^{-1}R_{2}^{2}R_{1}^{-1})R_{1}$ defining a geodesic in $\mathcal{P}_d$ connecting $R_{1}^2$ and $R_{2}^2$, and $U(t)$ defining the geodesic in $\mathcal{G}(d,n)$ given by Eq.~(\ref{eq:GeoGrass}). The distance $d_{\mathcal{S}^+}(G_1,G_2)$ between the two Gram matrices $G_1$ and $G_2$ results in the square of the length of the curve on $\mathcal{S}^+(d,n)$ 
\begin{equation}
\label{eq:closeness}
\begin{split}
d_{\mathcal{S}^+}(G_1,G_2)&=
d_{\mathcal{G}}^2(\mathcal{U}_{1},\mathcal{U}_{2})+kd_{\mathcal{P}_d}^2(R_{1}^2,R_{2}^2) \\
&=\|\Theta \|_F^2+k\|\log R_{1}^{-1}R_{2}^{2}R_{1}^{-1} \|_F^2.
\end{split}
\end{equation}
The parameter $k$ controls the influence of the squared Grassmann distance $d^2_{\mathcal{G}}$ and the squared Riemannian distance $d^2_{\mathcal{P}_d}$.
Figure ~\ref{fig:postive_cone} illustrates a graphical interpretation of this distance measurement. The detailed derivation is described in \cite{Bonnabel2009}. Some first application scenarios for this can be studied in \cite{Kacem2020}.

\subsection{Coherence Sequence Kernel}

Now consider the evolution of the shape geodesic  $\mathcal{C}_{G_(t-1)\to G_t}(t)$
as a stationary random process $X_t$. The frequency specific characteristics of the process can be described by a vector auto-regressive (VAR) model of order $p$ reflecting its coherence and causality \cite{Granger1969} written as
\begin{equation} \label{eq:var}
\centering
X_t = c + \sum_{i=1}^p \varphi_i X_{t-i}+ \varepsilon_t,
\end{equation}
with the intercept $c$, noise $\varepsilon$ and the coefficients $\varphi_i$ of the lags of $X$ till order $p$.
This model shares the properties of a bounded linear convolution operator on a normed sequence space 
\begin{equation}
\centering
x(t)=\int_{-\infty}^{\infty}\varphi(\tau)x(t-\tau) d\tau \
\end{equation}
defined with respect to the inner product of its underlying Hilbert space.
In case the Hilbert space is associated with a valid kernel that reproduces every function in the space one speaks of a reproducing kernel Hilbert space widely used in structural risk minimization as principals of machine learning \cite{Vapnik2000}.
Suppose any system\textquotesingle s impulse response $\Phi$ of the process in general can be represented as part of an universal prior
given by a multivariate Gaussian mixture model 

\begin{equation} \label{eq:ubm}
\centering
g(\Phi)= \sum_{c=1}^N \lambda_i\mathcal{M}(\Phi;\hat{\mu}_c,\hat{\sigma}_c).
\end{equation}
where $\lambda_c$ are the priors of the Gaussians $\mathcal{N}()$ and $\hat{\mu}_c$ and $\hat{\sigma}_c$
are the mean and covariance respectively. Given a set of observed responses $\tilde{\Phi} = \{\varphi_i \}$, $\mu$ can be estimated by the classical a posterior adaptation (MAP) given the universal prior $g(\Phi)$ \cite{Reynolds2000}. Each sample response $\varphi_j$ contributes to a particular Gaussian component $\mathcal{N}_c$ with respect to the posterior probability that $\varphi_i$ belongs to $\mathcal{N}_c$

\begin{equation}
\centering
r_i(c)=\frac{\mathcal{N}(\varphi_i;\hat{\mu}_c,\hat{\sigma}_c)}{\sum_{c=1}^M \mathcal{N}(\varphi_i;\hat{\mu}_c,\hat{\sigma}_c)}
\end{equation}
with the sufficient statistics
\begin{equation}
\centering
r_c=\sum_{i=1}^N r_i(c)
\end{equation}

\begin{equation}
\centering
z_c=\sum_{i=1}^N r_i(c)\varphi_i
\end{equation}
and the resulting MAP estimation
\begin{equation}
\centering
\mu_c=\frac{z_c+\frac{\sigma}{\hat{\sigma}}\hat{\mu}}{r_c+\frac{\sigma}{\hat{\sigma}}}
\end{equation}
This assumes identical priors and a diagonal covariance for the mixture components.\newline
The distance $d_{KL}(g(\Phi),g(\tilde{\Phi}))$ between the universal prior $g(\Phi)$ and the adapted distribution $g(\tilde{\Phi})$ can be expressed by a symmetric approximation of the Kullback-Leibler (KL) divergence \cite{Campbell2006}
\begin{equation}
\centering
d_{KL}(g(\Phi),g(\tilde{\Phi}))=\frac{1}{2} \sum_{c=1}^M \lambda_c(\hat{\mu}_c-\mu_c)\hat{\sigma}^{-1}_c(\hat{\mu}_c-\mu_c).
\end{equation}
The resulting inner product
\begin{equation} \label{eq:kernel}
\centering
K_{KL}(g(\Phi),g(\tilde{\Phi}))= \sum_{c=1}^M (\sqrt{\lambda_c}\hat{\sigma}^{-\frac{1}{2}}\hat{\mu})^t(\sqrt{\lambda_c}\hat{\sigma}^{-\frac{1}{2}}\mu)
\end{equation}
yields to a linear kernel function. The kernel is a simple scaled version of the vectorized means. A major advantage of such kind of kernel is that the dimension is constant across different observed sequence lengths.

\section{Experiments}
\label{sec:experiments}

\subsection{Data Collection}
\label{sec:database}

The longitudinal data was collected at the University Hospital, Aachen, Germany \footnote{This work involved human subjects in its research. Approval of all ethical and experimental procedures and protocols was granted by the Ethics Committee of the Medical Faculty of RWTH Aachen (Application No.: 356/19), and performed in line with the Declaration of Helsinki (World Medical Association, 2013 \cite{Ethics2013}). All participants of this study were financially reimbursed and gave their written informed consent for participation.}. An amount of 99 patients (52.3\% female, 47.7\% male, mean age 33.76 years, s.d. = 14.36) and 43 healthy controls (55.8\% female, 44.2\% male, mean age 37.03 years, s.d. = 14.88) were recruited from in-patient treatment facilities of the Department of Psychiatry, Psychotherapy and Psychosomatic, RWTH Aachen University. The patient group includes 66 participants with depression (65.2\% female, 34.8\% male, mean age 37.01 years, s.d. = 15.43)  and 33 with schizophrenia (39.4\% female, 60.6\% male, mean age 39.05 years, s.d. = 13.65). The first measurement took place immediately after they got hospitalized and the second short before their in-patient discharge. The time interval between these two measurements is 8 weeks on average. All participants of the control group were arbitrary recruited from the local population but interviewed and recorded with the same time interval between the measurements.

\subsubsection{Clinical Data}
\label{subsec:clinical_measures}
The clinical diagnoses were confirmed using the Structured Clinical Interview for DSM-5 Clinician Version (SCID-5-CV) and Structured Clinical Interview for DSM-5 Personality Disorders (SCID-5-PD) \cite{SCID5}. Symptom severity was assessed using the Brief Symptom Inventory (BSI) \cite{Derogatis1983}, the Beck Depression Inventory (BDI-II) \cite{Beck1961} and the GRID Hamilton Depression Rating Scale (GRID-HAMD 21) \cite{Williams2008}. A psychomotor vigilance task (PVT) battery was carried out assessing alertness, sustained attention and divided attention\cite{Dinges1985}. All participants of the control group were screened for a lifetime DSM-IV diagnosis (SCID-light) \cite{SCID5}, neurological illness or current substance misuse. 

\begin{figure*}[!h]
\begin{center}
\includegraphics[scale=.48]{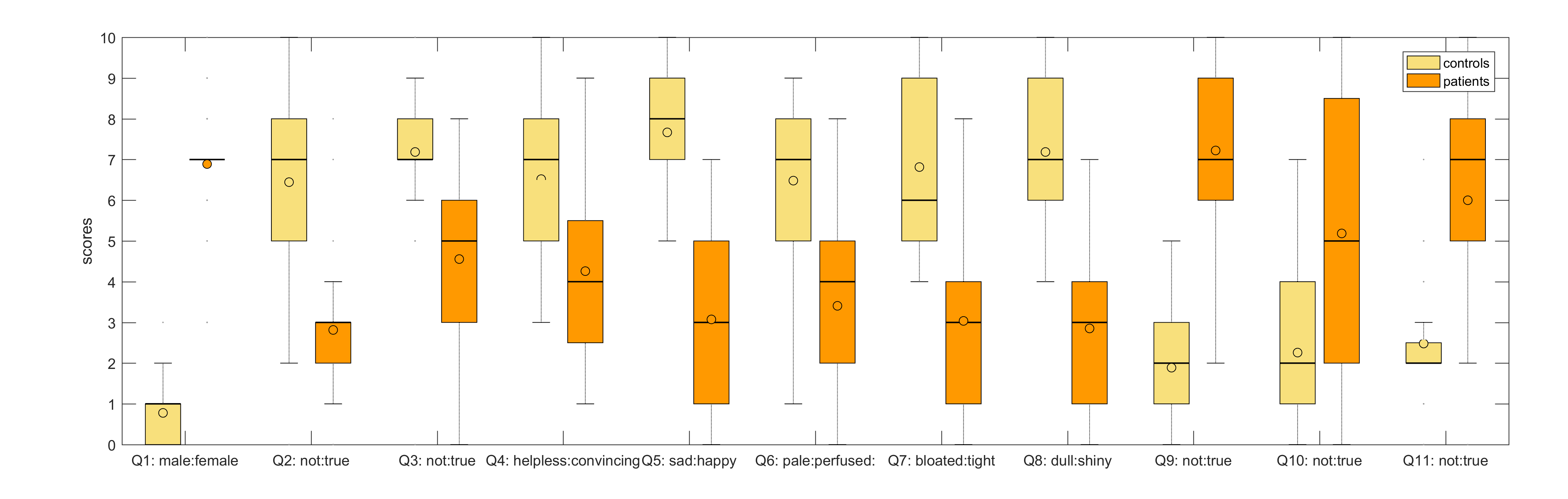}
\caption{The box plots for the mean face questionnaire. The individual questions are described in the list of subsection \ref{list:questinary}. The ratings of the answers vary between 0 and 10 and are plotted on the ordinate. The choice of answers is described on the abscissa.}
\label{fig:meanface_questionary}       
\end{center}
\end{figure*}

\subsubsection{Video Recordings}
\label{subsec:video_rec}
All patients were interviewed and simultaneously recorded with a standard consumer webcam (Logitech c270, 25 fps).   Each single measurement consists of three phases with an average recording duration of 90 minutes. Phase I includes the entire Hamilton interview between the participants and the clinican. During phase II all participants were shown videos of different facial expressions alternating with a neutral one which they were asked to imitate respectively. For every imitated mimic video they reported their own emotional state and intensity. The facial expression videos were taken from the Binghamton 3D Dynamic Facial Expression Database (BU-4DFE) \cite{BU4DFE}. The average duration of this phase is 10 minutes. In phase III all participants were presented with portrait-shot video clips of actors telling different short-stories. The general condition of the stories reflects emotionality in all channels while the other conditions either provided neutral speech content, facial expression, or prosody, respectively. Stimuli consist of 96 evaluated videos (duration, mean 10.93 s, s.d = 0.93), alternating a male (n = 44) or female (n = 52) protagonist who told a self-related story of disgusting, fearful, happy, sad or neutral situations. All actors had been instructed to imagine their story as vividly as possible, and to remember an emotionally corresponding life experience. The participants judged and reported the emotion and intensity presented, as well as their own emotional state and intensity \cite{Regenbogen2012}. The average duration of the third phase is 15 minutes.
All video recordings were manually checked for consistency. A manual segmentation of the measurement conditions was also carried out.

\subsubsection{Mean Faces}
\label{subsec:mean_faces}
In Figure \ref{fig:the_face_of_affective_disorders} the neutral mean faces of the patients and the healthy control group are visualized. The left face corresponds to the control group and the right face to the patient group. To calculate the mean faces a standard landmark fitting of the faces was carried out using the EmoNet framework \cite{toisoul2021estimation}. The selection of the neutral emotional face was done during phase II when the participants are instructed to imitate the specific facial expression. Procrustes analysis is applied to all fitted landmark shapes of the neutral face images. Piecewise affine transformation \cite{AAM2004} from the face shape to the mean face shape is determined in order to warp the face image to match the average shape and finally the mean for every pixel is computed.

\subsubsection{Feature Extraction}
\label{subsec:feature_extraction}

For each individual video recording landmark fitting, dimensional emotion recognition \cite{toisoul2021estimation} and eye gaze prediction was carried out \cite{Park2018}.
The shape geodesic (Eq. \ref{eq:geopsd}) was computed on a frame by frame basis and standard SVD based dimension reduction was applied. The resulting shape geodesic time series was concatenated with the rigid head pose and eye gaze vector and frame blocked with a windows size of ten seconds and an overlap of one second. Each block was modelled as VAR process of order three (Eq. \ref{eq:var}). The vectorized coefficient matrix $\varphi$ was further reduced in its dimension once again by utilizing SVD. Over the entire database, for all $\varphi$ an universal background model (Eq. \ref{eq:ubm}) was computed using the standard expectation maximization algorithm. MAP adaptation was performed for the different available segmentation phases of the recordings to obtain the final kernels as feature representation. In Figure \ref{fig:shape_geodesic_pipeline} the processing schematic is illustrated. For better reproducibility of the results, the individual dimensions of the vectors are shown here. VAR modeling and sequence kernel computation was also accomplished for the valence and arousal values of the EmoNet network \cite{toisoul2021estimation}.

\subsection{Statistical Analysis}
\label{sec:stats}

Various experiments were conducted to investigate fundamental issues related to the objectivity of current clinical diagnosis and treatment practices and how machine learning could contribute constructively in this area.  This was realized in the broadest sense via statistical evaluations and finally by applying a schematic of causal inference.
The first objective was to find out how affective disorders are generally perceived during social interaction and how accurate the clinical assessments really are. Secondly, we looked at how to automatically determine the most objective diagnosis possible as well as prognosis for treatment with the highest probability of success. The described model of facial dynamics serves as primary surrogate measurement for the underlying disturbed vigilance regulation and will be opposed to the discrete reaction time surrogate measurement of attention. The parameterized form of the facial dynamic is used as a predictor variable for several different clinical targets. In order to depict the respective statistical relationship, a regression model was trained using a standard Gaussian process with dot product covariance function over the sequence kernel (Eq. \ref{eq:kernel}).
The performance is measured utilizing a leave-one-subject-out cross validation over the entire database. Unless otherwise indicated, all results are presented in terms of the Pearson correlation coefficients obtained during the separate experiments. There was no violation of the statistical significance (e.g. p-values) in the obtained results, therefore this will be not stated explicitly further.

\subsubsection{Cues of Affective Disorders}
\label{subsec:cues}

The face contains a lot of information on which humans base their interactions with each other. Affective disorders may influence how others behave toward patients. To give a small impression how these people are perceived by their fellow human beings, 40 participants were surveyed to rate the mean faces with respect to several questions reflecting character traits and facial appearance. The participants were not informed that these faces are part of a clinical study including control subjects and patients. The following list shows the selected cues.

\begin{itemize}
\label{list:questinary}
  \item Q1: What is the gender of the two faces?
  \item Q2: Do the faces have an attractive appearance?
  \item Q3: Are these faces trustworthy persons?
  \item Q4: How do you assess the ability of these people to act?
  \item Q5: What is the emotion of the two faces?
  \item Q6/Q7: What is the skin appearance of the two faces?
  \item Q8: What is the impression of the gaze?
  \item Q9: Do the two faces have droopy mouth corners?
  \item Q10: Do the two faces have raised eye browns?
  \item Q11: Are these persons clinical patients?
\end{itemize}
In Figure \ref{fig:meanface_questionary} the box plots representing the distribution of the answers are visualized. These results show that the faces of individuals diagnosed with affective disorders were perceived as having more hanging mouth corners, raised eye browns, more swollen pale faces and a dull gaze. Affective disorders also were associated with female gender, sad emotion, less trustworthy, less leadership quality and less attractive face appearance. As a whole, this was also associated with illness.

\subsubsection{Clinical Data}
\label{subsec:clincial}

Apparently the mere static appearance of the facial expression already entails indications of a possible mental illness. The next logical question to be asked is which information the clinical interview contains. The common clinical interview is usually conducted in one session and lasts between 45 and 90 minutes. 
\begin{figure}[h!]
\begin{center}
\includegraphics[scale=.5]{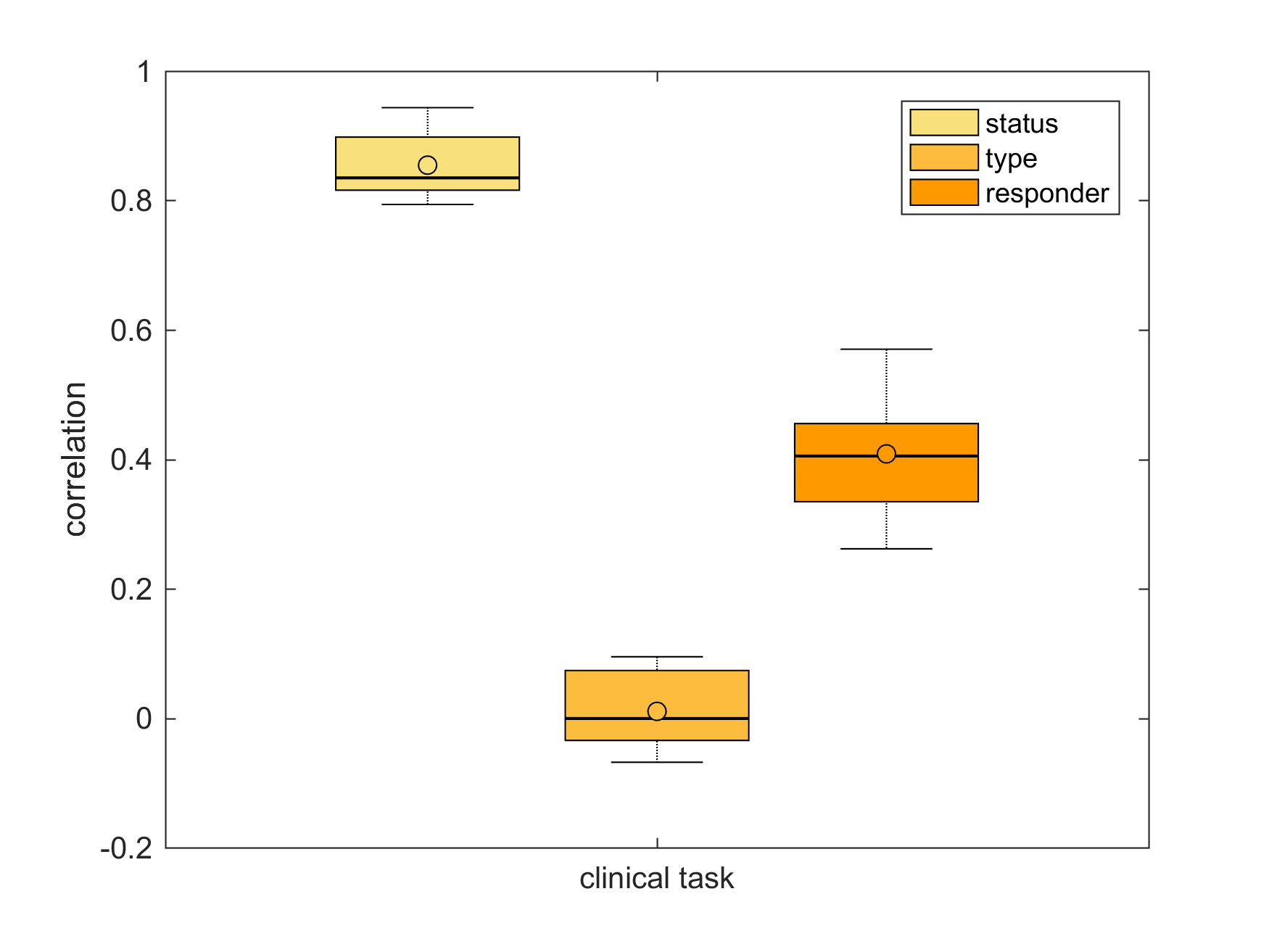}
\caption{The prediction of the patient status, type and treatment responder given the clinical symptom severity.}
\label{fig:clinical_results}       
\end{center}
\end{figure}
The duration varies depending on the complexity of the psychiatric history and can be up to three hours in complex cases.  In order to give a basic insight into the general problem of categorical clinical diagnostics, an attempt is made to determine the statistical predictive power using the results of the common observer based questionnaires. The symptom severity was used as concatenated feature vector including the BSI, the BDI and the HAMD scores. The target variable was selected as the overall patient status and patient type (e.q. depressive or schizophrenic) given by the SCID-5-CV assessment. As another target variable the treatment response given by an improvement of the symptom severity (HAMD solely) by at least 30 percent between the first and the second measurement was selected.
The results of the cross validation are shown in Figure \ref{fig:clinical_results}. The prediction of the patient status based upon the symptom severity features achieved an expected correlation of 0.82. However, the distinction between patient types only achieved an expected  correlation of -0.03. And the prediction of the treatment response achieved an expected correlation of 0.39. 
In essence, the patient status can be determined relatively well using the usual questionnaires. However, that is essentially all that can be concluded from it. Whether someone is depressed or rather schizophrenic is not indicated. The same applies to the treatment response. The results of the machine based interpretation and predictions utilizing the human facial behavior are described and presented in the following paragraphs.

\subsubsection{Patient Status}

\begin{figure}[h!]
\begin{center}
\includegraphics[scale=.5]{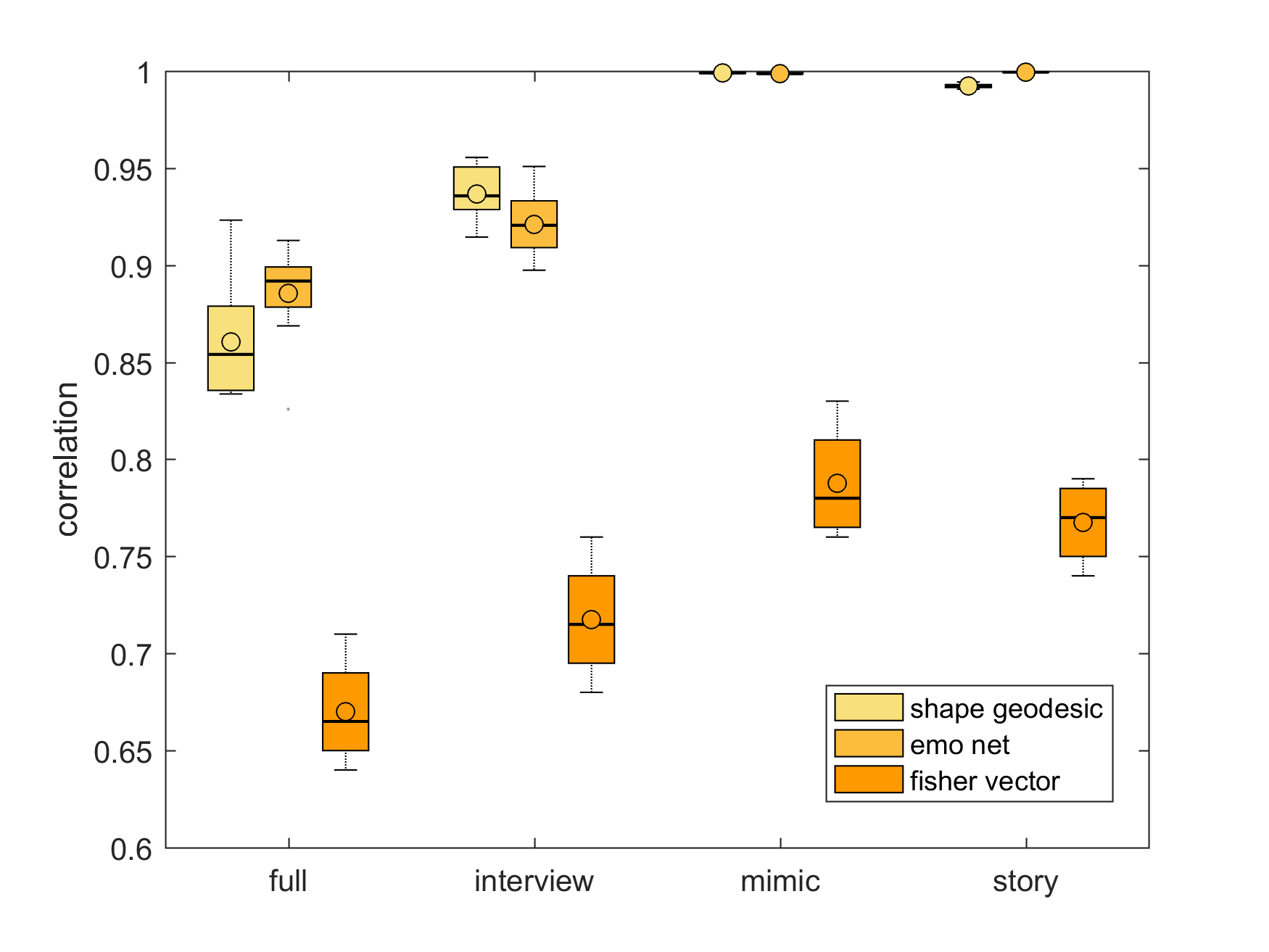}
\caption{The prediction of the patient status (e.q. healthy control or patient) given the coherence sequence kernel for the shape geodesic and the EmoNet valence-arousal time series compared to the Fisher vectors. The correlations are obtained at time of the hospitalization for the different measurement conditions (full, interview, mimic and story) respectively.}
\label{fig:shape_geodesic_vs_emoNet}       
\end{center}
\end{figure}

\begin{figure}[h!]
\begin{center}
\includegraphics[scale=.5]{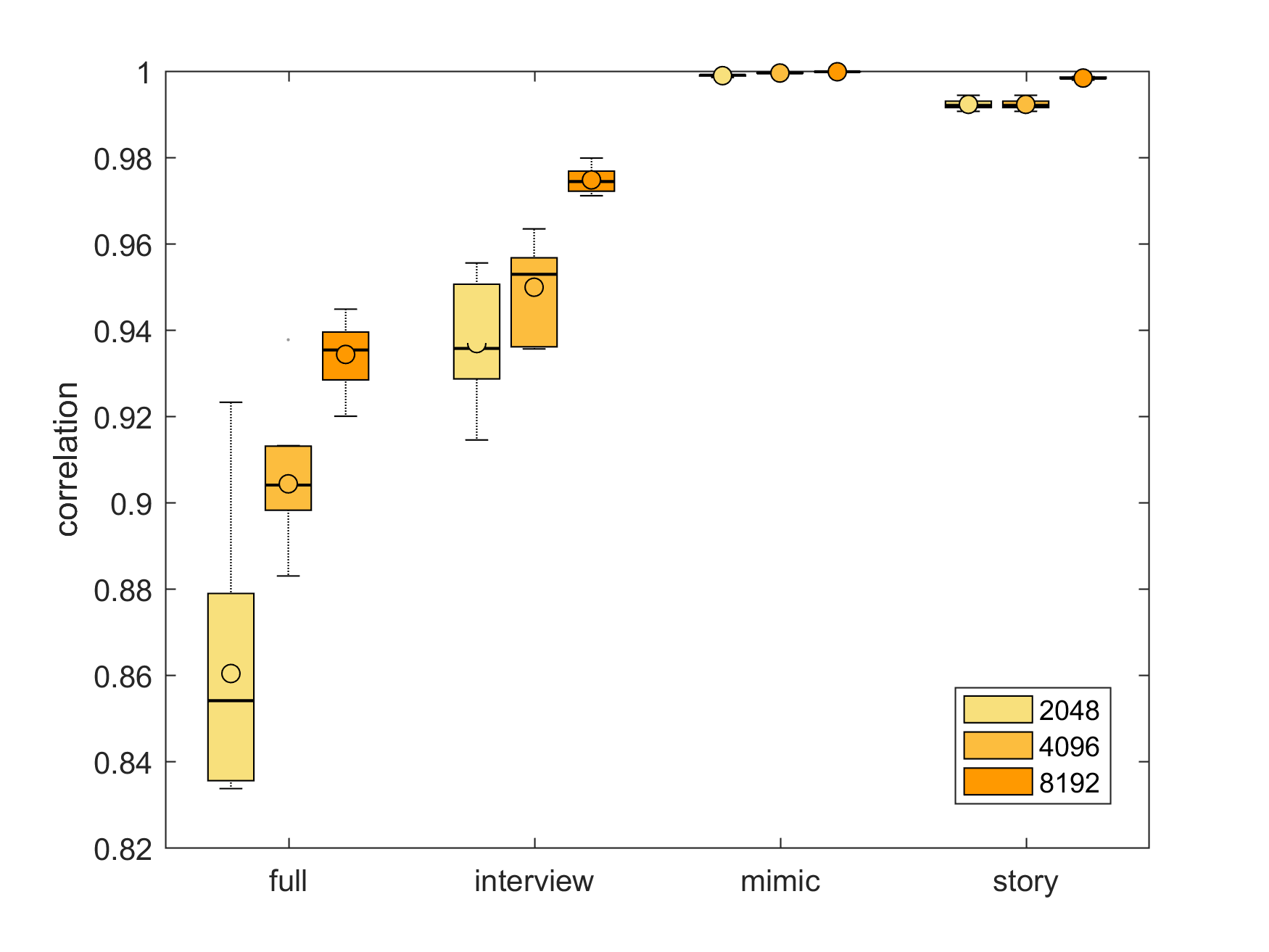}
\caption{The prediction of the patient status (e.q. healthy controls or patient) given the face dynamics at time of the hospitalization. The correlations are obtained for different mixture components of the background model (2048, 4096,
8192) and for the different measurement conditions (full, interview, mimic and story) respectively.}
\label{fig:patient_status_plot}       
\end{center}
\end{figure}

The first machine based prediction using the face dynamics
solely was conducted for the general patient status. To show a comparison, not only the proposed modeling of the face shape coherence was taken into account but also two other methods from this technical field.  The first alternative method is a deep learning approach based feature extraction methods (EmoNet) which projects the image based face regions to the Russel\textquotesingle s circumplex model of emotion \cite{Russel1980,toisoul2021estimation}. This dimensional representation is characterized by its valence and arousal values located on the unite circle.
\label{subsec:patientstatus}
\begin{figure}[h!]
\begin{center}
\includegraphics[scale=.6]{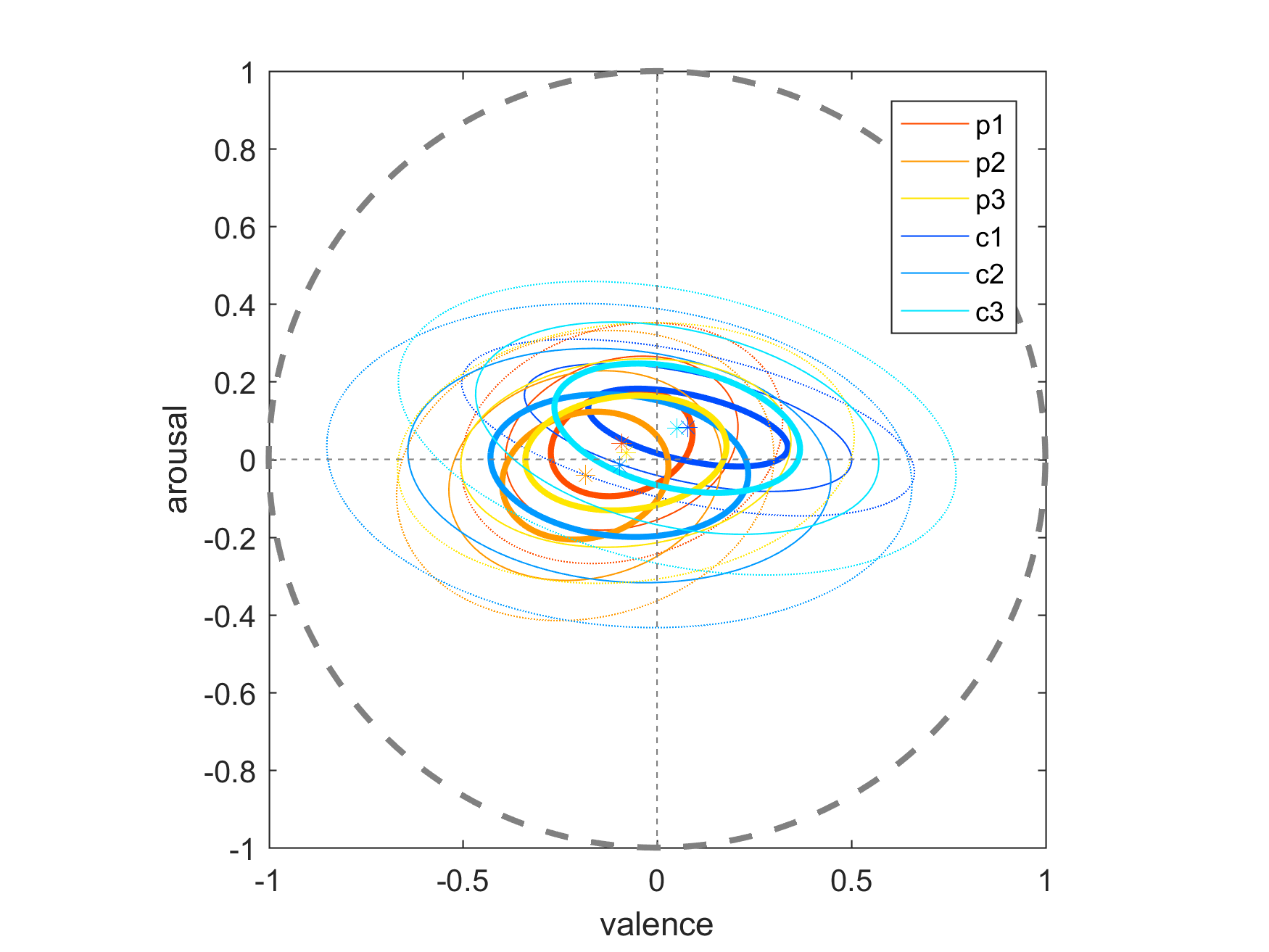}
\caption{The valence and arousal distributions predicted by the EmoNet over the three different measurement conditions (1,2,3) for both groups, the patients (p) and the controls (c).}
\label{fig:va_circumplex}       
\end{center}
\end{figure}
Figure \ref{fig:va_circumplex} visualizes the valence and arousal distributions by its mean and covariance computed over the three different measurement conditions for the patients  and controls group. The yellow tone ellipses (p1, p2 and p3) correspond to the patients group and the blue tone ellipses (c1, c2 and c3) to the controls group. The patients show lower arousal and lower valence expectations. This means patients are less activated and are showing a more negative mood. Overall, the controls group has a much higher dynamic range of valence which reflects a more pronounced regulation of their emotional behavior. The valence and arousal distributions represent a static point of view of the facial behavior. Nevertheless, the patients already show a difference here compared to the controls, analogous to the snapshot of the presented mean faces (see Figure \ref{fig:the_face_of_affective_disorders}).
In the next step VAR modeling and sequence kernels were computed for the shape geodesic as well as for the valence and arousal time series. As second alternative method a fisher vector encoding is computed over the barycentric representation of the face shape \cite{Kacem2018}. The feature computation for all three methods was accomplished once for the entire measurement and then for each of the three individual phases of the measurements (in the following Figures labeled by \textit{full}, \textit{interview}, \textit{mimic} and \textit{story}).
For each method, each set of features and the patient status as the target, the cross-validation for the Gaussian process regression was executed. The number of components for the mixture models is set to 2048 for both sequence kernel variants and set to 16 for the fisher vectors. The comparison of scored correlations is represented in Figure \ref{fig:shape_geodesic_vs_emoNet}.
For the complete measurement the prediction of the patient status based upon the coherence sequence kernel (labeled as shape geodesic) achieved an expected correlation of 0.85, the EmoNet kernel an expected correlation of 0.89 and the fisher vector an expected correlation of 0.67. For the interview phase the coherence sequence kernel achieved an expected correlation of 0.93, the EmoNet kernel an expected correlation of 0.92 and the fisher vector an expected correlation of 0.72. For the mimic phase the coherence sequence kernel achieved an expected correlation of 0.99, the EmoNet kernel an expected correlation of 0.98 and the fisher vector an expected correlation of 0.77. For the story phase the coherence sequence kernel achieved an expected correlation of 0.991, the EmoNet kernel an expected correlation of 0.998 and the fisher vector an expected correlation of 0.76.
Both sequence kernels show a show similar performance which is almost flawless. The fisher vector shows a slightly weaker performance. However, it must be mentioned here that the EmoNet network has to be trained on the basis of around 450,000 manually annotated face images in order to be able to predict emotions at all. In a further evaluation the scalability of the coherence sequence kernel was examined. For this purpose, the complexity of the Gaussian components was increased. The comparison of scored correlations for different numbers of mixture components is visualized in Figure \ref{fig:patient_status_plot}.
\begin{figure}[h!]
\begin{center}
\includegraphics[scale=.5]{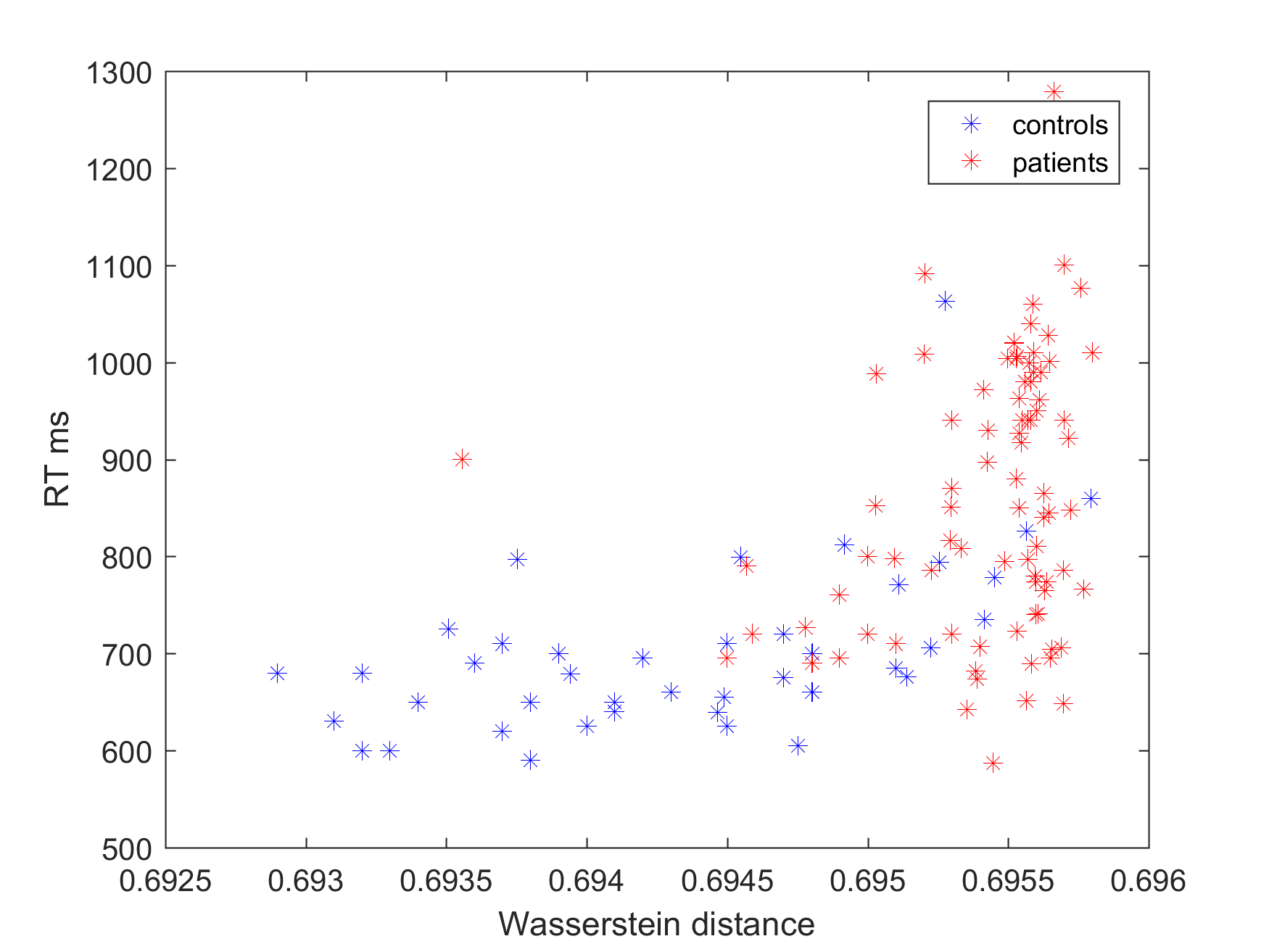}
\caption{The reaction times (RT) of the attention test versus the Wasserstein distances of the adapted priors probabilities with respect to the priors of the background mixture model.}
\label{fig:prior_rt_plot}       
\end{center}
\end{figure}
It is clear that predictive performance increases as components become more complex. Full saturation of performance is evident in the mimic phase. To emphasize this explicitly, the mimic phase shows the best performing correlations overall. And the measuring time is approximately 10 minutes which is relatively short compared to the rather lengthy clinical measurements. In addition, the distance between the adapted prior probabilities and the prior probabilities of background mixture model was calculated by the Wasserstein distance. This measure of the frequency of specific regulation of facial dynamics is compared to the reaction times of the attention test. In Figure \ref{fig:prior_rt_plot} this comparison is shown for both groups, patients and controls. The larger the distances become, the more likely it is to be able to observe a longer reaction time. Reaction times are often associated with severity of symptoms and are also linked to vigilance regulation \cite{Kloesch2022}.
\subsubsection{Patient Type}
\label{subsec:patienttype}
\begin{figure}[h!]
\begin{center}
\includegraphics[scale=.5]{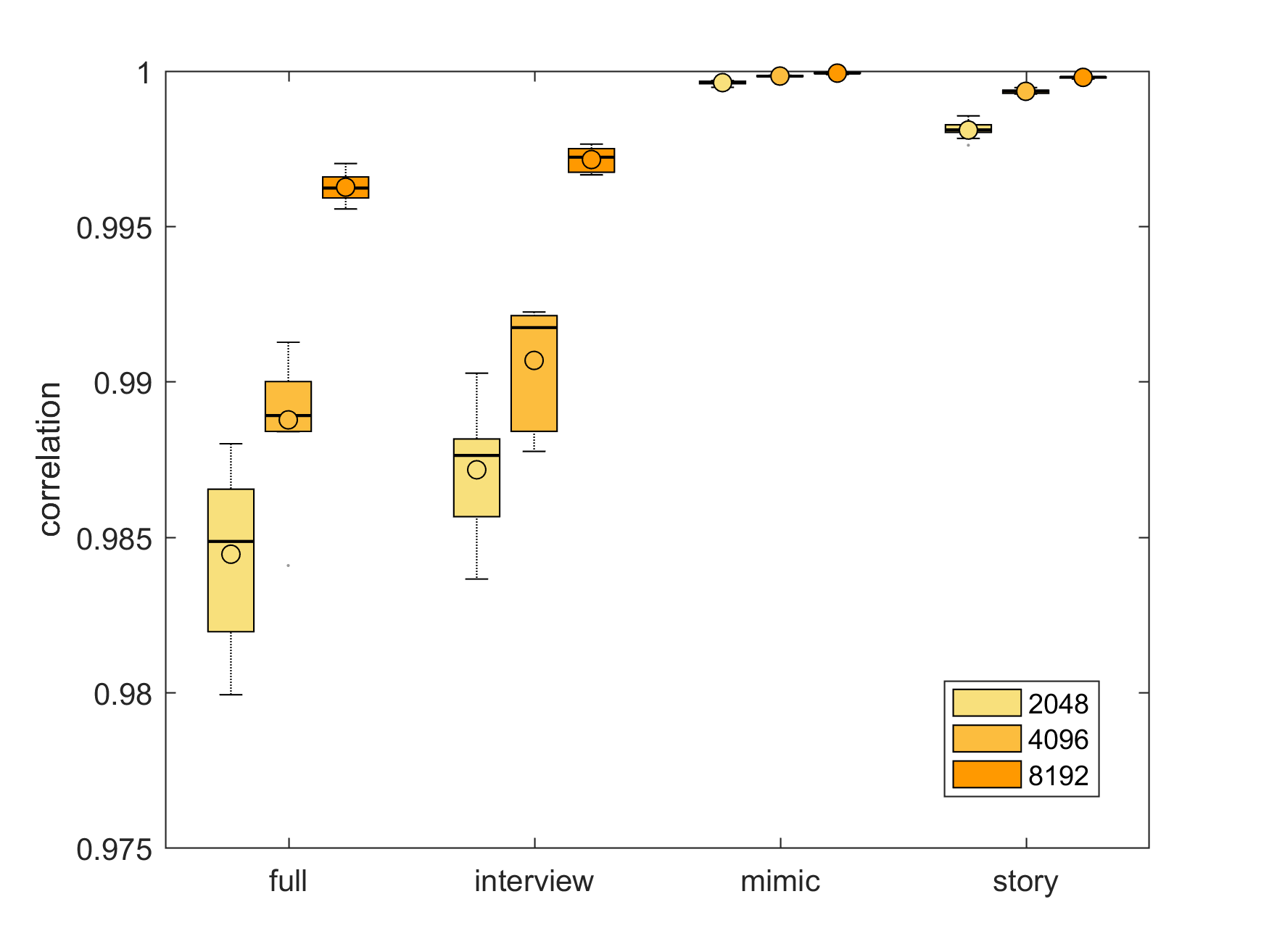}
\caption{The prediction of the patient type (e.q. depressive or schizophrenic) given the face dynamics at time of the hospitalization. The correlations are obtained for
different mixture components of the background model (2048, 4096,
8192) and for the different measurement conditions (full, interview, mimic and story) respectively.}
\label{fig:patient_intra_plot}       
\end{center}
\end{figure}

Studies have shown that a plain static observation of facial expressions does not distinguish between depressive and schizophrenic patients \cite{Gaebel2004}. However, on the other hand it could already be demonstrated by means of a rather rough observer based assessment of non-verbal behaviors that the schizophrenics differed by a decreased number of head movements, the depressive by a decreased eyebrow raising and both patient groups, but especially the depressed, by an overall reduction of the facial expressions associated with laughter \cite{Jones1979}. Now the question is whether a dynamic embedding of the facial expressions yields to a machine based distinction in this case. The procedure for this is the same as for estimating the patient status. Only the target variable is different here and set to the corresponding patient type. The cross-validation was only carried out over the entire patient group. Only the coherence sequence kernel was used as a feature vector, with different Gaussian components in order to check the scalability. The comparison of the calculated correlations is visualized in Figure  \ref{fig:patient_intra_plot}.
The scalability is very similar to the previous estimation of the patient status and reaches almost complete saturation for the mimic phase of the measurements again. Apart from that, all measurement conditions achieved a correlation of more than 0.98, which shows an almost error-free estimation of patient type.

\subsubsection{Treatment Responder}
\label{subsec:treatment}
A pharmacological treatment corresponds to a large extent to the standard procedure for patients during their inpatient stay. One of the aims is to initially reduce the acute symptoms, at least temporarily, so that other therapeutic measures can be carried out at all. However, this is also often the method of choice for longer therapeutic intervention. The local protein synthesis of the neurotransmitters is a very complex process. The interaction of the individual components has not been fully clarified to the greatest possible extent in order to always be able to achieve an optimal effect in the treatment of patients. The availability of active substances is also limited. The phenomenon of non-response is therefore one of the main problems when selecting the pharmacological agent to be used. Prevalence estimates of treatment resistance vary from 20 to 60 percent of all patients \cite{Howes2022}. As already pointed out in sub-chapter 7.2, the purely empirical clinical observer ratings including BSI, BDI and HAMD as predictior variables can only be used to very vaguely estimate response to drug treatment. 
It is now appropriate to specify the influence that machine-interpreted facial dynamics may have on the prediction of therapy response.
First, it should be evaluated if the treatment response can be predicted generally. Next, whether an individual drug treatment recommendation by means of the sensor-based monitoring of the face behavior can lead to an objective and more successful approach tailored to the patients needs.

\subsubsection*{Clinical}
\label{paragraph:clinical}

\begin{figure}[h!]
\begin{center}
\includegraphics[scale=.5]{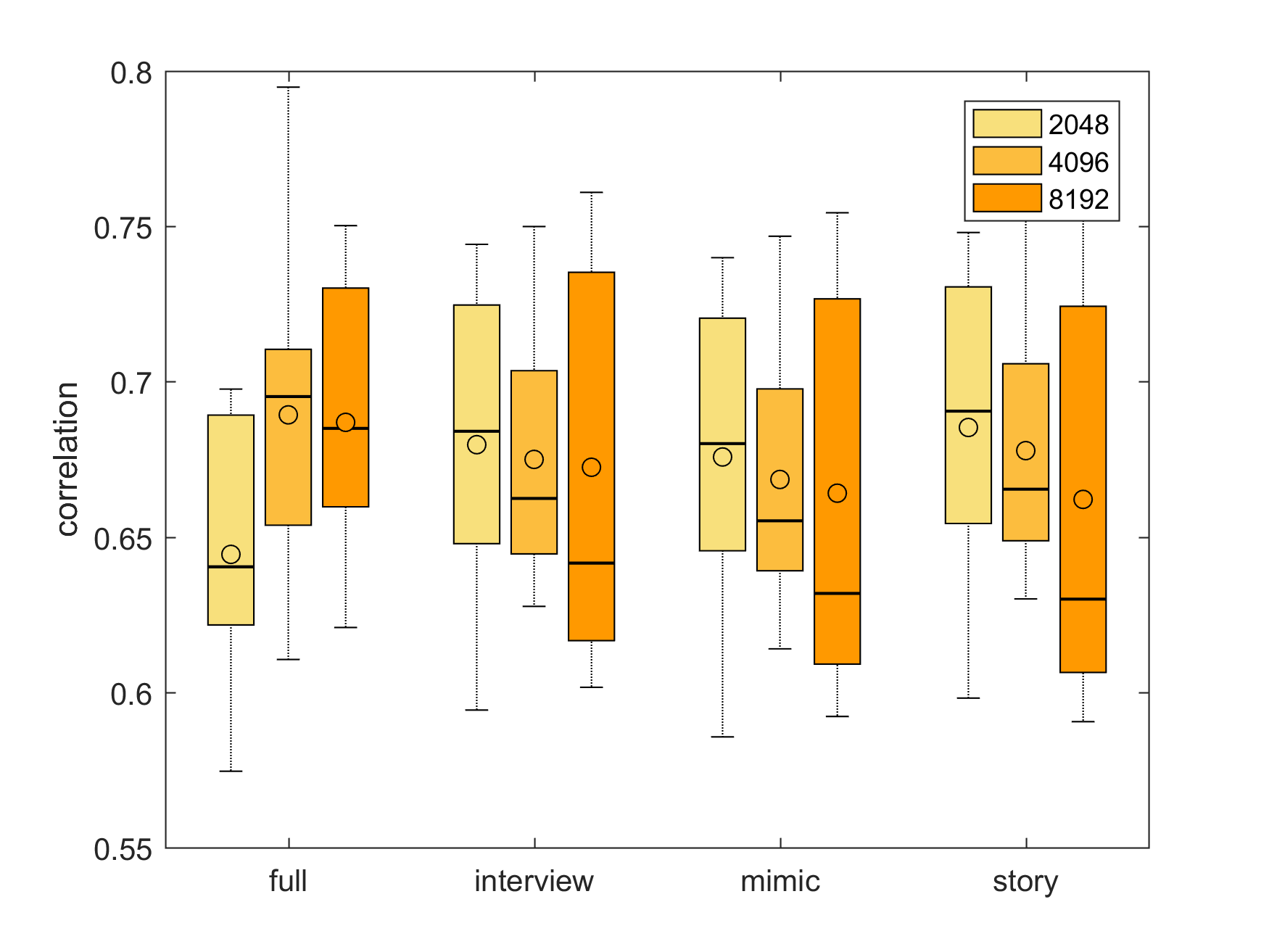}
\caption{The prediction of the symptom severity (HAMD-17) for the in-patient discharge measurements given the face dynamics at time of the hospitalization. The correlations are obtained for
different mixture components of the background model (2048, 4096,
8192) and for the different measurement conditions (full, interview, mimic and story) respectively.}
\label{fig:responder_dim}       
\end{center}
\end{figure}

To estimate the treatment response the HAMD-17 is used as target variable for the cross-validation of the Gaussian process regression. Consequently, it was evaluated here whether the facial dynamics monitored during the inpatient admission can provide information about the possible severity of symptoms at the future time of discharge. For a given symptom severity at the beginning of the treatment, the responder behavior can be derived. The corresponding  comparison of the calculated correlations is shown in Figure  \ref{fig:responder_dim}. Again the scalability was determined by the model complexity, varying the number of mixture components for the background model. 
\begin{figure}[h!]
\begin{center}
\includegraphics[scale=.5]{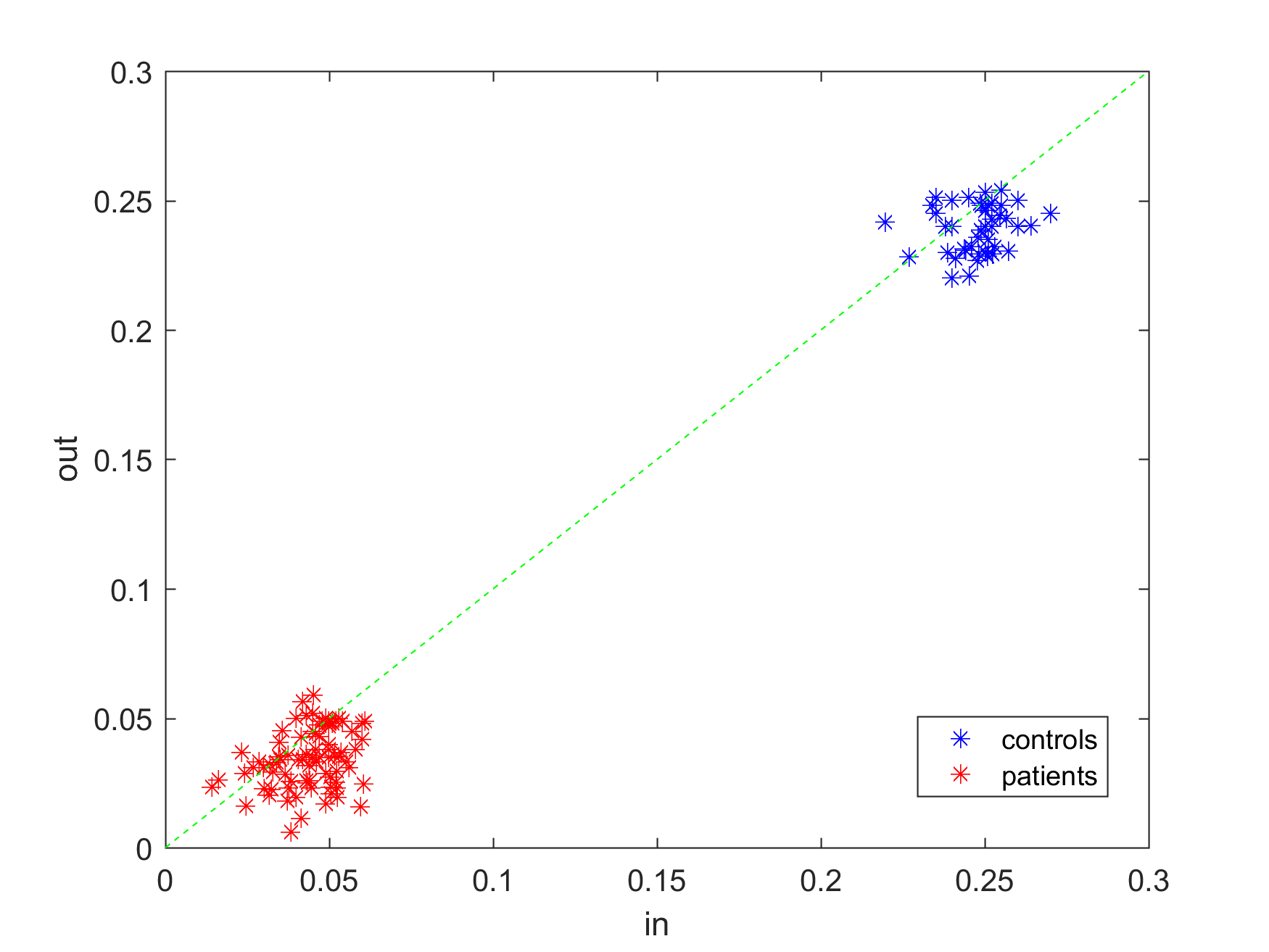}
\caption{The dot products of the sequence kernel with respect to the mean sequence kernel of the control group. The distances are plotted for the hospitalization measurements (in) and the in-patient discharge (out) for both groups, the patients and the controls.}
\label{fig:dot_product}       
\end{center}
\end{figure}
What immediately becomes apparent is that scalability has been lost.
The expectation of the correlations ranges between 0.63 and 0.69 over the different measurement conditions. This is noticeably less compared to the two previous evaluations. At this point one would assume that if the clinical assessment changes or the symptoms improve, then the face dynamics as a surrogate measurement of the underlying vigilance regulation should change accordingly.

This can be checked very easily by calculating the dot product of the feature vectors with respect to the mean vector computed over the control group, once for the in-patient
admission (in) and another time for the discharge (out). In
Figure \ref{fig:dot_product} the dot products for the patients (red) and the control
group (blue) are visualized. As expected the distances for the
controls tend not to change. However, this is also the case with
the patients. Wouldn\textquotesingle t one logically expect the distances to 
move towards the healthy ones? It is remarkable that the
reaction times of the attention test also show a similar
development.
\subsubsection*{Causal Inference}
\label{paragraph:Causal}
\begin{figure}[h!]
\begin{center}
\includegraphics[scale=.5]{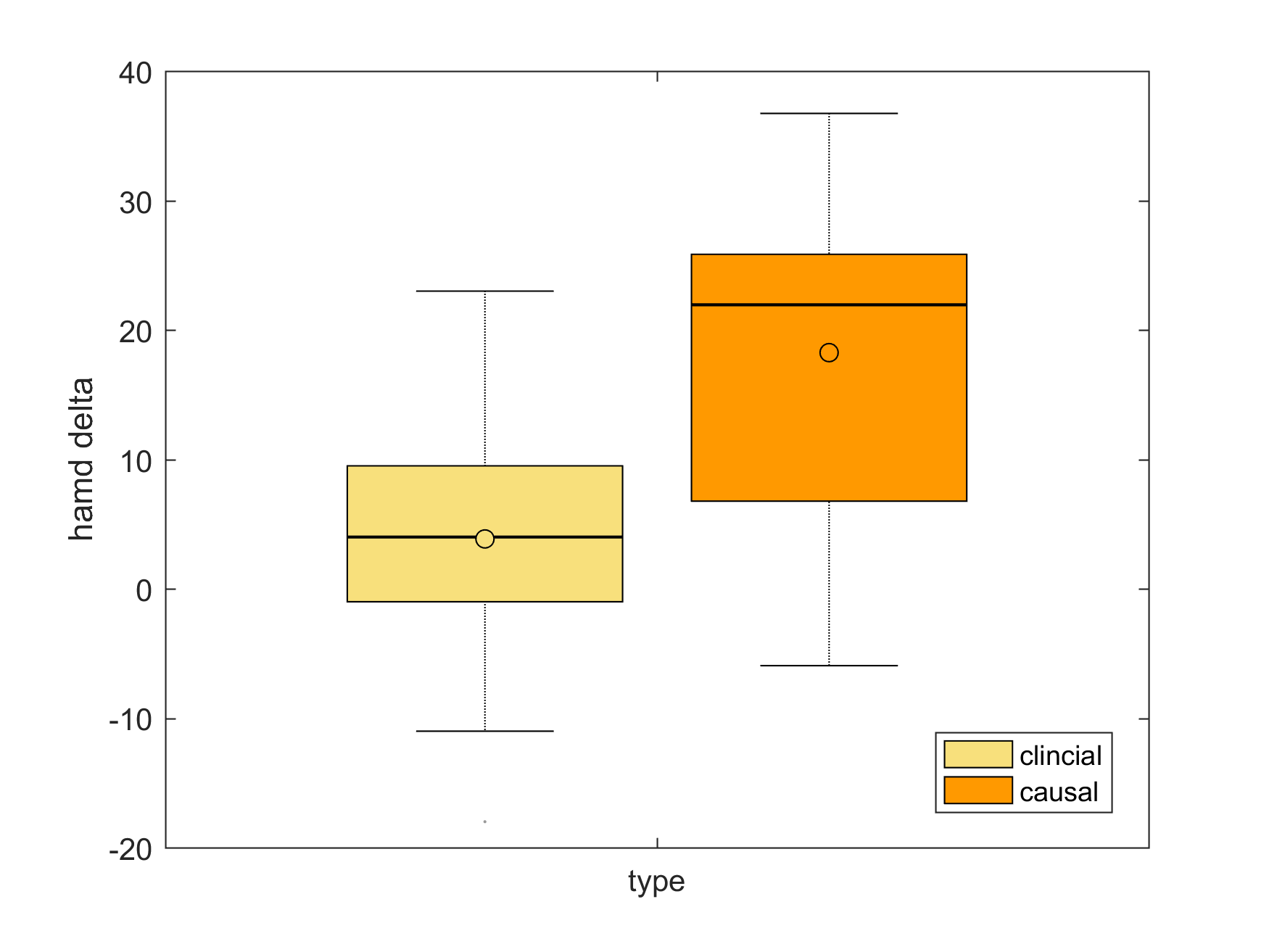}
\caption{The difference in symptom severity (HAMD-17 delta) between time of the hospitalization and the in-patient discharge for the classical clinical assessment (clinical) and the computed causal (causal) obtained ones.}
\label{fig:tensor_causal}       
\end{center}
\end{figure}

\begin{figure}[h!]
\begin{center}
\includegraphics[scale=.5]{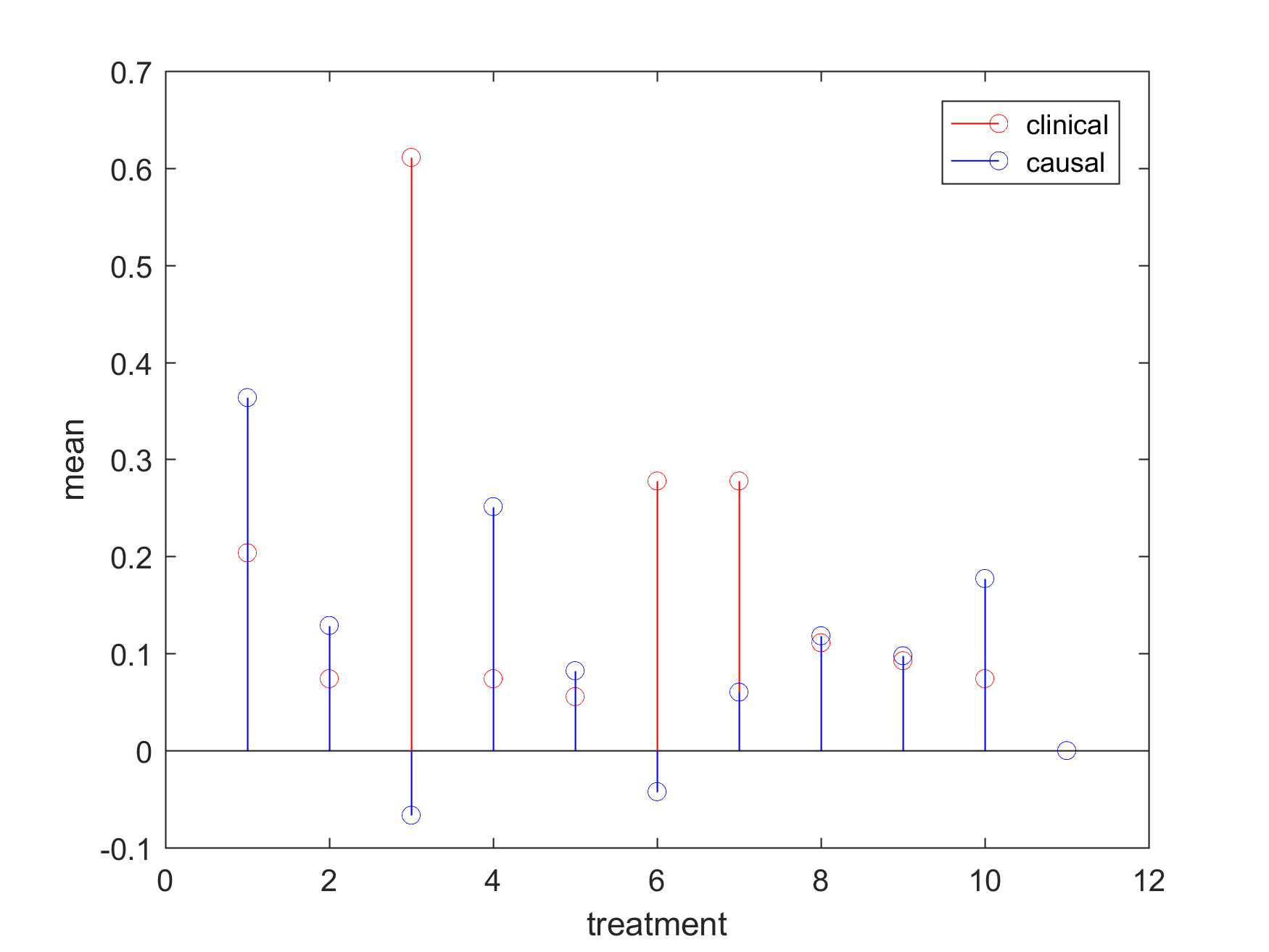}
\caption{The comparison of the average prescribed pharmacological treatment between the clinical assessment (red) and the predicted causal recommendation (blue). The pharmacological treatment types are listed in Table \ref{pharmacological_treatments}.}
\label{fig:causal_treatment}       
\end{center}
\end{figure}

\begin{figure}[h!]
\begin{center}
\includegraphics[scale=.5]{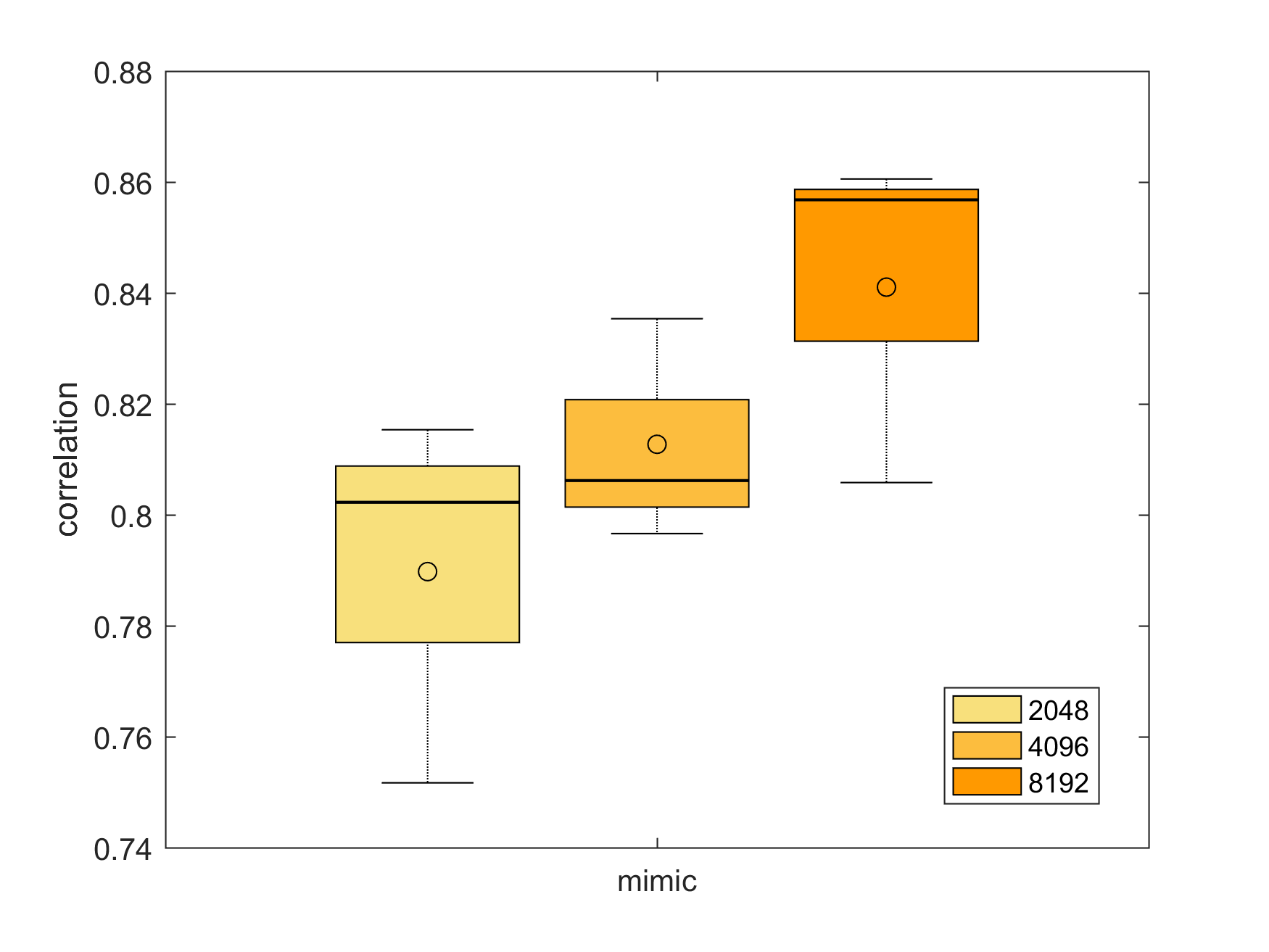}
\caption{The prediction of the symptom severity (HAMD-17) for the in-patient discharge measurements given the face dynamics at time of the hospitalization and the causal treatment recommendation. The correlations are obtained for
different mixture components of the background model (2048, 4096,
8192) and for the mimic measurement condition.}
\label{fig:responder_prediction_causal}       
\end{center}
\end{figure}

\begin{figure}[h!]
\begin{center}
\includegraphics[scale=.5]{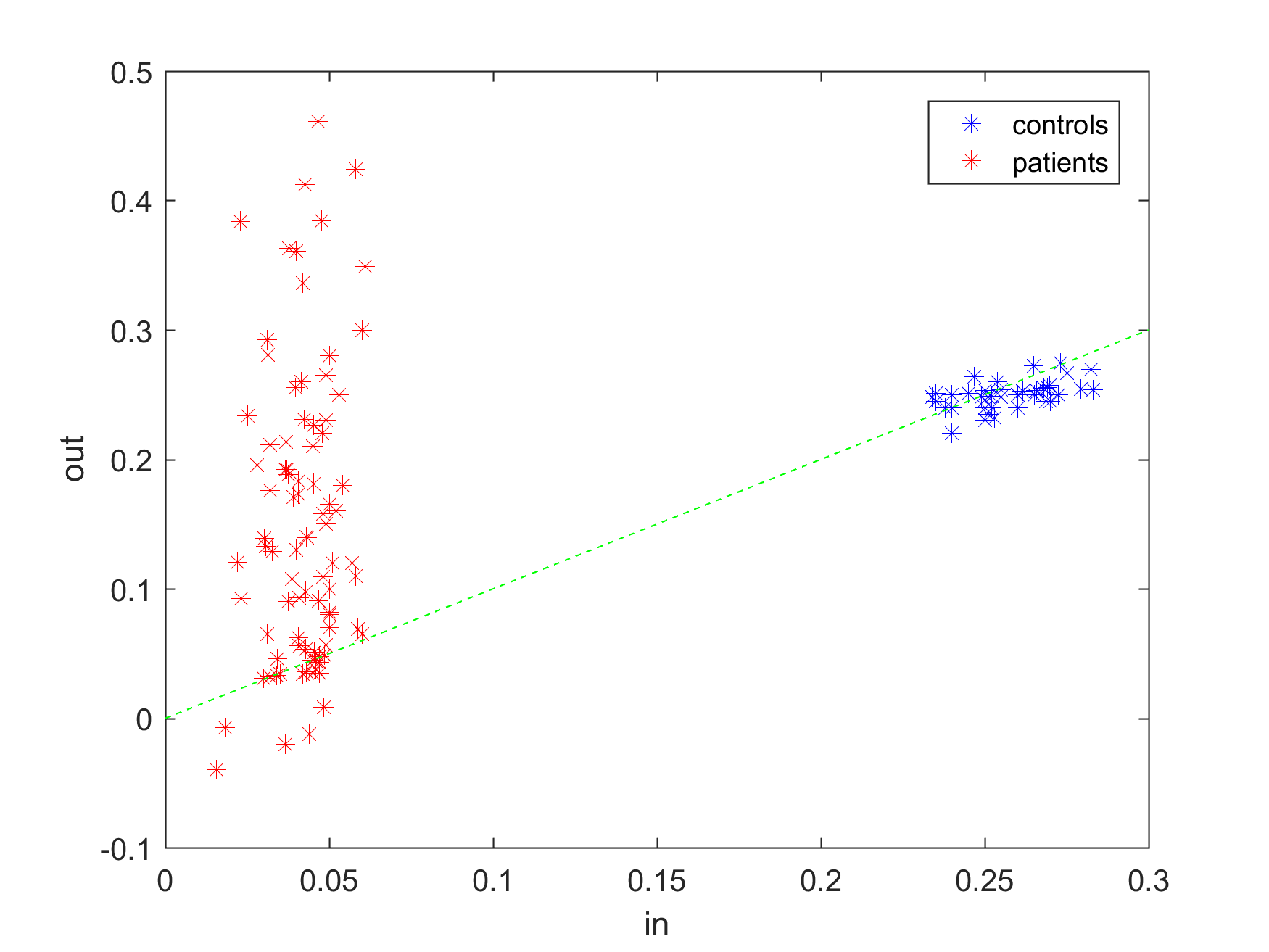}
\caption{The dot products of the causal predicted sequence kernel with respect to the mean sequence kernel of the control group. The distances are plotted for the hospitalization measurements (in) and the in-patient discharge (out) for both groups, the patients and the controls.}
\label{fig:causal_dot}       
\end{center}
\end{figure}

The last experiment evaluates the performance of the face dynamics as a linear model of causal inference for the treatment selection.
Recalling causality in the original sense is questioning cause and effect in order to justify action.
In intelligence, causality is more about questioning options for action.
Put simply, we ask the question: what if? Consequently, a counterfactual.
We put a condition on a possible action and check what would happen based on empirical experience \cite{Pearl2018}.
\begin{figure}[h!]
\begin{center}
\includegraphics[scale=.24]{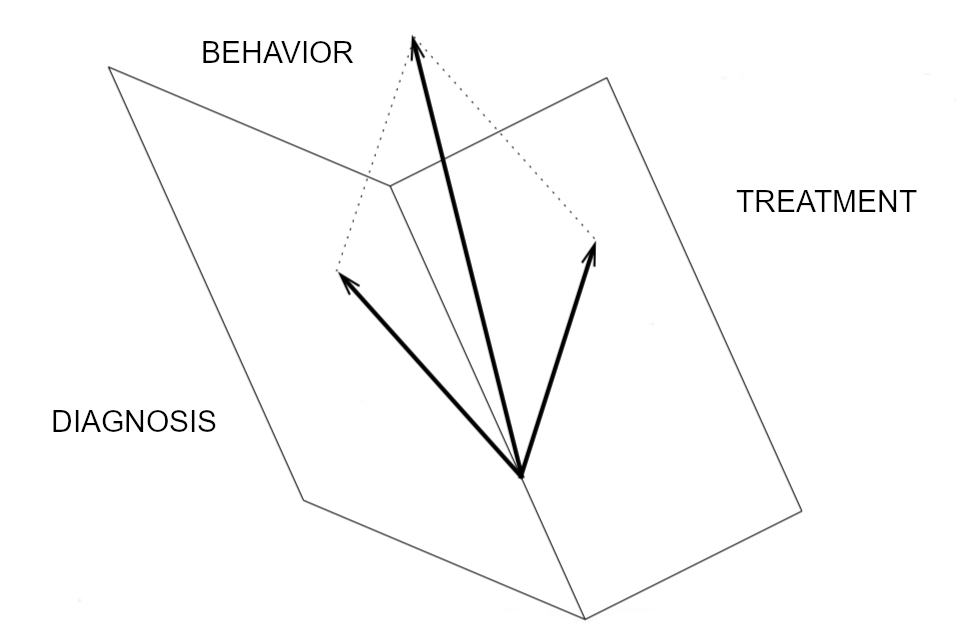}
\caption{The measurement space including the face behavior, the clinical diagnosis and the treatment as a schematic vector space. Each mode of the measurements is associated with its own subspace.}
\label{fig:tensor}       
\end{center}
\end{figure}
The empirical prior knowledge is available in the form of the behavioural face dynamics ($U_1$), the medication as treatment ($U_2$) and the diagnosis with its symptom severity ($U_3$).
In trivial terms, all components of the behavioural face pattern can now be tested to see if they trend positively towards the behavioural patterns of the healthy samples for a given pharmacological treatment option. We thus obtain a probability, the best choice for a given observation.
Technically, we form a tensor and calculate the subspace for each mode as a model of linearly independent basis vectors \cite{Vasilescu2002} 
\begin{equation}
\centering
\mathcal{D}=\mathcal{Z} \times_1 U_1 \times_2 U_2 \times_3 U_3
\end{equation} 
with $\mathcal{Z}$ the core tensor and $U_1$, $U_2$ and $U_3$ the corresponding mode matrices. For a given projection of the face dynamics, we alter the treatment preserving the identity of the face dynamics until a clinically relevant improvement in depressive symptonmatology, assess via the HAMD, reached. This can be expressed and solved as a standard constrained optimization problem. We defined a reduction of the HAMD by at least 50 percent as constrained for the model. Figure \ref{fig:tensor} illustrates the measurements as a schematic vector space. In Table \ref{pharmacological_treatments} the pharmacological treatment types prescribed during the study are listed. In Figure \ref{fig:tensor_causal} the difference in symptom severity (HAMD delta) for the classical clinical assessment and the computed causal obtained ones is visualized. The expected improvement of symptom severity during the clinical treatment (clinical) is a reduction of the HAMD score of four points. For the synthetic altered treatment (causal) based upon the linear causal inference model the expected improvement of symptom severity is a reduced HAMD score of 22 points in the mean. 
\begin{table}[!t]
\renewcommand{\arraystretch}{1.3}
\caption{The pharmacological treatment types prescribed during the study.}
\label{pharmacological_treatments}
\centering
\begin{tabular}{|c||c|}
\hline
ID & Category\\
\hline
1 & Antipsychotica\\
2 & Antikonvulsiva/-epileptica\\
3 & Anxiolytica (upon need)\\
4 & Anxiolytica (daily)\\
5 & Opiode\\
6 & SSRI\\
7 & SNRI\\
8 & SNDRI\\
9 & Tetracyclic AD\\
10 & Tricyclic AD \\
11 & MAO\\
\hline
\end{tabular}
\end{table}
In Figure \ref{fig:causal_treatment} the difference of the average prescribed treatment per category between the clinical assessment and the predicted causal recommendation is opposed.
It can be observed that the causal model completely rejects the prescription of anxiolytika as needed and instead selected daily usage, if at all. Also the model rejects the prescription of
the reuptake-inhibitors (SSRI and SNRI). Instead, the model proposed antipsychotics and tricyclic antidepressants. Using the facial dynamics achieved through the synthetic changed treatment, a cross-validation was used for estimating the treatment response once again. For this evaluation only the mimic phase of the video measurements was examined. However, value was placed on the scalability by increasing the number of components of the background model. In Figure \ref{fig:responder_prediction_causal} the scored correlations are visualized. For the model with 2048 mixture components the expected correlation is 0.801. The model with 4096 mixture components 
achieved an expected correlation of 0.805. The model with 8192 mixture components achieved the highest expected correlation with 0.857. The scalability is given here again. 
Compared to the clinical estimate, a noticeably better estimate can be observed here. Finally, for the synthesized treatment, the dot products over both measurement times was calculated.
Figure \ref{fig:causal_dot} shows these for the patients and the controls. It should be noted here that the patient distances are now moving towards the controls. That would mean, at the time of discharge, there is a noticeable reduction in the affective flattening. In essence a return to a physiological regulation of the vigilance.

\subsubsection*{Limitations}
The results of the model are, in part, in contrast to standard, guideline-based clinical routines. 
Especially with respect to anxiolytics and SSRI, indicating that further research is needed here to 
investigate specific clinical indications and the overall prognostic utility of the present causal inference model
in greater detail. Interpretations of the data presented here can only be made with strong caution,
since the model incorporates the aspect of treatment only in the form of pharmacological therapy. 
However treatment of affective disorders for the patients in this sample entailed additional, 
multidimensional treatment options, including different forms of psychotherapy, 
non-invasive brain stimulation, electroconvulsive therapy, light therapy, 
occupational therapy and physical therapy. 
Hence, conclusions or recommendations for clinical practice cannot be made based solely on the data presented here.

\section{Future Work}
Future studies in this domain will help to recognize additional factors, such as gender or comorbidity, that further contribute to the enormous complexity of psychiatric disorders and the therapeutic mechanisms. While our results might be preliminary, they exemplify the huge potential of incorporating advanced machine-learning techniques and predictive modelling in psychiatric research via an interdisciplinary approach. Treatment response is often very limited in psychiatry, and more personalized approaches, enabling treatment prediction at the single-subject level are
urgently needed.  These automatically collected and processed data might offer a cost-effective means to generate clinically relevant, personalized insights and predictions, thereby capturing novel digital markers of behavior. If confirmed in future studies, causal inference models such as the one presented here will make such predictions more personalized with respect to potential outcomes and
disease course.

\section{Conclusion}
We have studied the human facial appearance and behavior in the clinical field of affective disorders and introduced a coherence sequence kernel based on the theory of hemifacial asymmetry. We were able to show that facial dynamics are affected when clinically diagnosed as a state which can be linked to rigid regulation of vigilance. This shows a pathological behavior which not only differs significantly from healthy people, but is also clear between the two mental disorders depression and schizophrenia. We have further demonstrated that the change in facial dynamics over a longer period of time contains information about the response to pharmacological treatment. Overall, the results predicted by the machine show better correlations compared to the pure clinical observer rating based questionnaires and are also objective. The relatively short measurement period of a few minutes for the computer vision approaches is also noteworthy, whereas hours are sometimes required for the clinical interviews or questionnaires. However, it is important to emphasize that any human diagnosis involves much more modalities and also involves very associative skills. For this reason, it is still too early to speak of a better general diagnosis option through machine vision at this point. The causal inference over the empirical prior knowledge of the data collection adjusted the pharmacological treatment in order observe a return to the physiological regulation of the facial dynamics. Such a return couldn\textquotesingle t be observed during the clinical prescription. At the moment it is not clear whether such a machine based recommendation would indeed result to a significant better success of therapy. Further studies, incorporating well controlled, interventional study designs as well as larger patient samples, are needed to corroborate the present results. However, such kind of patient-tailored approaches would break the barriers of the common categorical classification schematic still dominantly used in daily life. We hope to report about a successful alignment of the presented methodology in one of our future works. Nevertheless, this still requires some further research before the method we named \textit{Opto-Electronic Encephalography (OEG)} will be fully suitable for everyday use.


%



\ifCLASSOPTIONcompsoc
  \section*{Acknowledgments}
\else
  \section*{Acknowledgment}
\fi

This work was funded, in part, by the European Regional Development Fund (EFRE): LifeScience.NRW under grant agreement EFRE LS-2-2-030c and by CanControls GmbH Aachen/ Germany.

\ifCLASSOPTIONcaptionsoff
  \newpage
\fi



%



\bibliographystyle{IEEEtran}
\bibliography{IEEEabrv,egbib}

\begin{thebibliography}{10}
\providecommand{\url}[1]{#1}
\csname url@samestyle\endcsname
\providecommand{\newblock}{\relax}
\providecommand{\bibinfo}[2]{#2}
\providecommand{\BIBentrySTDinterwordspacing}{\spaceskip=0pt\relax}
\providecommand{\BIBentryALTinterwordstretchfactor}{4}
\providecommand{\BIBentryALTinterwordspacing}{\spaceskip=\fontdimen2\font plus
\BIBentryALTinterwordstretchfactor\fontdimen3\font minus
  \fontdimen4\font\relax}
\providecommand{\BIBforeignlanguage}[2]{{%
\expandafter\ifx\csname l@#1\endcsname\relax
\typeout{** WARNING: IEEEtran.bst: No hyphenation pattern has been}%
\typeout{** loaded for the language `#1'. Using the pattern for}%
\typeout{** the default language instead.}%
\else
\language=\csname l@#1\endcsname
\fi
#2}}
\providecommand{\BIBdecl}{\relax}
\BIBdecl

\bibitem{Lancet2022}
G.~. M.~D. Collaborators*, ``Global, regional, and national burden of 12 mental
  disorders in 204 countries and territories, 1990–2019: a systematic
  analysis for the global burden of disease study 2019.'' \emph{The Lancet
  Psychiatry}, vol.~9, pp. 137–--150, 2022.

\bibitem{Cuthbert2013}
B.~N. Cuthbert and T.~R. Insel, ``Toward the future of psychiatric diagnosis:
  the seven pillars of rdoc.'' \emph{BMC Med}, vol.~11, no. 126, 2013.

\bibitem{Haro2014}
J.~H. et~al., ``Roamer: roadmap for mental health research in europe.''
  \emph{Int J Methods Psychiatr Res}, vol.~1, pp. 1--14, 2014.

\bibitem{Yuste2014}
R.~Y. G.~M. Chruch, ``The new century of the brain.'' \emph{Sci Am}, vol. 310,
  no.~3, pp. 38--45, 2014.

\bibitem{Wykes2015}
T.~W. et~al., ``Mental health research priorities for europe.'' \emph{The
  Lancet Psychatry}, vol.~2, no.~1, pp. 1036--1042, 2015.

\bibitem{Topol2014}
E.~J. Topol, ``Individualized medicine from prewomb to tomb.'' \emph{Cell},
  vol. 157, no.~1, pp. 241--253, 2014.

\bibitem{Bzdok2018}
D.~Bzdok and A.~Meyer-Lindenberg, ``Machine learning for precision psychiatry:
  opportunities and challenges.'' \emph{Biol Psychiatry Cogn Neurosci
  Neuroimaging}, pp. 223--230, 2018.

\bibitem{Hahn2017}
T.~H. et~al., ``Predictive analytics in mental health: applications,
  guidelines, challenges and perspectives.'' \emph{Mol Psychiatry}, vol.~22,
  no.~1, pp. 37--43, 2017.

\bibitem{Shamout2021}
F.~Shamout, T.~Zhu, and D.~A. Clifton, ``Machine learning for clinical outcome
  prediction.'' \emph{IEEE Reviews in Biomedical Engineering}, vol.~14, pp.
  116–--126, 2021.

\bibitem{Ulrich2013}
G.~Ulrich, ``The theoretical interpretation of electroencephalography (eeg).''
  \emph{BMED Press LLC}, 2013.

\bibitem{Bente1964}
D.~Bente, ``Die {I}nsuffizienz des {V}igilit\"astonus.''
  \emph{Habilitationsschrift}, 1964.

\bibitem{Hegerl2014}
U.~Hegerl and S.~Olbrich, ``The vigilance regulation model of affective
  disorders and adhd.'' \emph{Neuroscience and Biobehavioral Reviews}, vol.~44,
  pp. 45--57, 2014.

\bibitem{Pilz2020}
C.~S. Pilz, I.~B. Makhlouf, U.~Habel, and S.~Leonhardt, ``Predicting brainwaves
  from face videos.'' \emph{The IEEE Conference on Computer Vision and Pattern
  Recognition (CVPR) Workshops}, pp. 1139--1147, 2020.

\bibitem{Kacem2018}
A.~Kacem, Z.~Hammal, M.~Daoudi, and J.~Cohn, ``Detecting depression severity by
  interpretable representations of motion dynamics.'' \emph{The IEEE
  International Conference on Automatic Face and Gesture Recognition (FG
  2018)}, pp. 739–--745, 2018.

\bibitem{Harati2020}
S.~Harati, A.~Crowell, Y.~Huang, H.~Mayberg, and S.~Nemati, ``Classifying
  depression severity in recovery from major depressive disorder via dynamic
  facial features.'' \emph{IEEE Journal of Biomedical and Health Informatics},
  vol.~24, no.~3, pp. 815–--824, 2020.

\bibitem{Dukes2021}
D.~Dukes, K.~Abrams, and R.~A. et~al., ``The rise of affectivism.''
  \emph{Nature Human Behaviour}, vol.~5, pp. 816--820, 2021.

\bibitem{Granger1969}
C.~Granger, ``Investigating causal relations by econometric models and
  cross-spectral methods.'' \emph{Econometrica}, vol.~37, no.~3, pp. 424--438,
  1969.

\bibitem{Mandal1998}
H.~S. Asthana and M.~K. Mandal, ``Hemifacial asymmetry in emotion
  expressions.'' \emph{Behav Modif}, vol.~22, pp. 177–--183, 1998.

\bibitem{Aristoteles1913}
Aristoteles, ``Physiognomica.'' \emph{In W.D. Ross (Ed.) and T. Loveday and
  E.S. Forster (Transl.). The works of Aristotle. Oxford, England: Clarendon},
  pp. 805–--813, 1913.

\bibitem{Darwin1872}
C.~Darwin, ``The expression of the emotions in man and animals.'' \emph{John
  Murray, London}, 1872.

\bibitem{Werner1982}
H.~D. Werner, ``Cognitive therapy: A humanistic approach.'' \emph{Free Press},
  1982.

\bibitem{Eibl_Eibesfeldt1975}
I.~Eibl-Eibesfeldt, ``Die {B}iologie des menschlichen {V}erhaltens.''
  \emph{Piper, München}, 1975.

\bibitem{Tomkins1962}
S.~S. Tomkins, ``Affect, imagery, consciousness, vol. 1: The positive
  affects.'' \emph{Springer, New York}, 1962.

\bibitem{Izard1977}
C.~E. Izzard, ``Human emotions.'' \emph{Plenum Press, New York}, 1977.

\bibitem{Ekman1992_2}
P.~Ekman, ``An argument for basic emotions.'' \emph{Cogn Emot}, vol.~6, pp.
  169–--200, 1992.

\bibitem{Jack2014}
R.~E. Jack, O.~Garrod, and P.~G. Schyns, ``Dynamic facial expressions of
  emotion transmit an evolving hierarchy of signals over time.'' \emph{Current
  Biology}, vol.~24, no.~2, pp. 187–--195, 2014.

\bibitem{Cannon1929}
W.~B. Cannon, ``Bodily changes in pain, hunger, fear and rage.''
  \emph{Appleton}, 1929.

\bibitem{Bracha2004}
S.~Bracha, ``Freeze, flight, fight, fright, faint: Adaptationist perspectives
  on the acute stress response spectrum.'' \emph{CNS Spectrums}, pp. 679--685,
  2004.

\bibitem{Leonhard1968}
K.~Leonhard, ``Der menschliche ausdruck.'' \emph{Johann Ambrosius Barth,
  Leipzig}, 1968.

\bibitem{Bleuler1911}
E.~Bleuler, ``{L}ehrbuch der {P}sychiatrie.'' \emph{Springer, Heidelberg},
  1911.

\bibitem{Pideret1896}
T.~Pideret, ``{M}imik und {P}hysiognomik.'' \emph{Meyersche Hofbuchhandlung,
  Detmold}, 1896.

\bibitem{Grabowski2019}
K.~Grabowski, A.~Rynkiewicz, A.~Lassalle, S.~Baron-Cohen, B.~Schuller,
  N.~Cummins, A.~Baird, J.~Podgorska-Bednarz, A.~Pieniazek, and I.~Lucka,
  ``Emotional expression in psychiatric conditions: New technology for
  clinicians.'' \emph{Psychiatry Clin Neurosci}, vol.~73, no.~2, pp. 50--62,
  2019.

\bibitem{Song2022}
S.~Song, S.~Jaiswal, L.~Shen, and M.~Valstar, ``Spectral representation of
  behaviour primitives for depression analysis.'' \emph{IEEE Transactions on
  Affective Computing}, vol.~13, no.~2, pp. 829--844, 2022.

\bibitem{Schroedinger1944}
E.~Schr\"odinger, ``{W}hat is {L}ife? {T}he {P}hysical {A}spect of the {L}iving
  {C}ell.'' \emph{Cambridge University Press}, 1944.

\bibitem{MaturanaVarela1973}
H.~Maturana and F.~Varela, ``Autopoiesis and cognition: The realization of the
  living.'' \emph{Boston Studies in the Philosophy of Science}, vol.~42, 1973.

\bibitem{Prigogine1977}
I.~Prigogine, ``Time, structure and fluctuations.'' \emph{Nobel Lecture, 8.
  Dezember 1977}, 1977.

\bibitem{Shannon1948_2}
C.~Shannon, ``A mathematical theory of communication.'' \emph{Bell System
  Technical Journal}, vol.~27, no.~3, pp. 379–--423, 1948.

\bibitem{Hurrelmann1989}
K.~Hurrelmann, ``Human development and health.'' \emph{Springer, New York}, pp.
  5–--6, 1989.

\bibitem{Head1923}
H.~Head, ``The conception of nervous and mental energy - vigilance: a
  physiological state of the nervous system.'' \emph{British Journal of
  Psychology}, vol.~14, pp. 126--147, 1923.

\bibitem{Bateson1979}
G.~Bateson, ``Mind and nature: A necessary unity.'' \emph{New York: E. P.
  Dutton}, 1979.

\bibitem{Carnap1956}
R.~Carnap, ``The methodological character of theoretical concepts.'' \emph{The
  Foundations of Science and the Concepts of Psychology and Psychoanalysis,
  University of Minnesota Press}, pp. 38--76, 1956.

\bibitem{Ulrich1988}
G.~Ulrich, ``Addendum to: The importance of the concept of vigilance for
  psychophysiological research. {H}eld at {U}niversity of {L}eipzig 2008,''
  \emph{Medical Hypotheses}, vol.~27, pp. 227--229, 1988.

\bibitem{Davis1938}
H.~Davis, P.~A. Davis, A.~L. Loomis, E.~N. Harvey, and G.~Hobart, ``Changes in
  human brain potentials during the onset of sleep.'' \emph{Journal of
  Experimental Psychology}, vol.~1, pp. 24--38, 1938.

\bibitem{Loomis1937}
A.~l.~Loomis, E.~N. Harvey, and G.~A. Hobart, ``Celebral states during sleep as
  studied by human brain potentials.'' \emph{Journal of Experimental
  Psychology}, vol.~21, pp. 127--144, 1937.

\bibitem{Roth961}
B.~Roth, ``The clinical and theoretical importance of eeg rhythms corresponding
  to states of lowered vigilance.'' \emph{Clinical Neurophysiology}, vol.~13,
  pp. 395--399, 1961.

\bibitem{Ulrich1986}
G.~Ulrich and K.~Frick, ``A new quantitative approach to the assessment of
  stages of vigilance as defined by spatiotemporal eeg patterning.''
  \emph{Perceptual and Motor Skills}, vol.~62, no.~2, pp. 567--576, 1986.

\bibitem{Kloesch2022}
G.~Kl\"osch, J.~Zeitlhofer, and O.~Ipsiroglu, ``Revisiting the concept of
  vigilance.'' \emph{Front Psychiatry}, 2022.

\bibitem{Kendall1989}
D.~G. Kendall, ``A survey of the statistical theory of shape.'' \emph{Statist
  Sci}, vol.~4, no.~2, pp. 87--99, 1993.

\bibitem{Gower1982}
J.~Gower, ``Euclidean distance geometry.'' \emph{Mathematical Scientist},
  vol.~7, pp. 7--14, 1982.

\bibitem{Begelfor2006}
E.~Begelfor and M.~Werman, ``Afﬁne invariance revisited.'' \emph{IEEE
  Computer Society Conference on Computer Vision and Pattern Recognition}, pp.
  2087--2094, 2006.

\bibitem{Bonnabel2009}
S.~Bonnabel and R.~Sepulchre, ``Riemannian metric and geometric mean for
  positive semideﬁnite matrices of ﬁxed rank.'' \emph{SIAM Journal on
  Matrix Analysis and Applications}, vol.~31, no.~3, pp. 1055--1070, 2009.

\bibitem{Kacem2020}
A.~Kacem, M.~Daoudi, B.~Amor, S.~Berreti, and J.~Alvarez-Paiva, ``A novel
  geometric framework on gram matrix trajectories for human behavior
  understanding.'' \emph{IEEE Transactions on Pattern Analysis and Machine
  Intelligence}, vol.~42, no.~1, pp. 1--14, 2020.

\bibitem{Vapnik2000}
V.~Vapnik, ``The nature of statistical learning theory.'' \emph{Information
  Science and Statistics. Springer-Verlag}, 2000.

\bibitem{Reynolds2000}
D.~A. Reynolds, T.~F. Quatieri, and R.~B. Dunn, ``Speaker verification using
  adapted gaussian mixture models.'' \emph{Digital Signal Processing}, vol.~10,
  no.~1, pp. 19--41, 2000.

\bibitem{Campbell2006}
W.~M. Campbell, D.~E. Sturim, and D.~A. Reynolds, ``Support vector machines
  using gmm supervectors for speaker verification.'' \emph{IEEE Signal
  Processing Letters}, vol.~13, no.~5, pp. 308–--311, 2006.

\bibitem{Ethics2013}
``World medical association declaration of helsinki: ethical principles for
  medical research involving human subjects.'' \emph{JAMA}, vol. 310, no.~20,
  pp. 2191--2194, 2013.

\bibitem{SCID5}
``Structured clinical interview for dsm-5 (scid-5).'' \emph{American
  Psychiatric Association Publishing}, 2017.

\bibitem{Derogatis1983}
L.~R. D.~N. Melisaratos, ``The brief symptom inventory: an introductory
  report.'' \emph{Psychological medicine}, vol.~13, no.~3, pp. 595–--605,
  1983.

\bibitem{Beck1961}
A.~T. Beck, C.~H. Ward, M.~Mendelson, J.~Mocj, and J.~Erbaugh, ``An inventory
  for measuring depression.'' \emph{Archives of General Psychiatry}, vol.~4,
  pp. 561--571, 1961.

\bibitem{Williams2008}
J.~B.~W. et~al., ``The grid-hamd: standardization of the hamilton depression
  rating scale.'' \emph{International clinical psychopharmacology}, vol.~23,
  no.~3, pp. 120–--129, 2008.

\bibitem{Dinges1985}
I.~D. Dinges and J.~W. Powell, ``Microcomputer analysis of performance on a
  portable, simple visual reaction time task sustained operations.''
  \emph{Behavior Research Methods, Instrumentation, and Computers}, vol.~17,
  pp. 652--655, 1985.

\bibitem{BU4DFE}
L.~Yin, X.~Wei, Y.~Sun, J.~Wang, and M.~Rosato, ``A 3d facial expression
  database for facial behavior research.'' \emph{The IEEE Conference on
  Automatic Face and Gesture Recognition}, pp. 211--216, 2006.

\bibitem{Regenbogen2012}
C.~Regenbogen, D.~A. Schneider, A.~Finkelmeyer, N.~Kohn, B.~Derntl,
  T.~Kellermann, R.~E. Gur, F.~Schneider, and U.~Habel, ``The differential
  contribution of facial expressions, prosody, and speech content to empathy.''
  \emph{Cognition and Emotion}, vol.~26, no.~6, pp. 995--1014, 2012.

\bibitem{toisoul2021estimation}
A.~Toisoul, J.~Kossaifi, A.~Bulat, G.~Tzimiropoulos, and M.~Pantic,
  ``Estimation of continuous valence and arousal levels from faces in
  naturalistic conditions,'' \emph{Nature Machine Intelligence}, 2021.

\bibitem{AAM2004}
I.~Matthews and S.~Baker, ``Active appearance models revisited.''
  \emph{International Journal of Computer Vision}, pp. 135--164, 2004.

\bibitem{Park2018}
S.~Park, X.~Zhang, A.~Bulling, and O.~Hilliges, ``Learning to find eye region
  landmarks for remote gaze estimation in unconstrained settings.'' \emph{In
  Proceedings of the 2018 ACM Symposium on Eye Tracking Research and
  Applications}, pp. 21--27, 2018.

\bibitem{Russel1980}
J.~Russell, ``A circumplex model of affect.'' \emph{Journal of Personality and
  Social Psychology}, vol.~39, no.~6, pp. 1161–--1178, 1980.

\bibitem{Gaebel2004}
W.~Gaebel and W.~Woelwer, ``Facial expressivity in the course of schizophrenia
  and depression.'' \emph{Archives of Psychiatry and Clinical Neurosciences},
  vol. 254, pp. 335–--342, 2004.

\bibitem{Jones1979}
I.~Jones and M.~Pansa, ``Some nonverbal aspects of depression and schizophrenia
  occurring during the interview.'' \emph{The Journal of Nervous and Mental
  Disease}, vol.~16, no.~7, pp. 402–--409, 1979.

\bibitem{Howes2022}
O.~D. Howes, M.~E. Thas, and T.~Pillinger, ``Treatment resistance in
  psychiatry: state of the art and new directions.'' \emph{Mol Psychiatry},
  vol.~27, pp. 58–--72, 2022.

\bibitem{Pearl2018}
J.~Pearl and D.~Mackenzie, ``The book of why: The new science of cause and
  effect.'' \emph{Basic Books}, 2018.

\bibitem{Vasilescu2002}
M.~A.~O. Vasilescu and D.~Terzopoulos, ``Multilinear analysis of image
  ensembles: Tensorfaces.'' \emph{Proc 7th European Conference on Computer
  Vision, Springer-Verlag, Berlin}, vol. 2350, no.~3, pp. 447--460, 2002.

\end{thebibliography}

%

\newpage

\begin{IEEEbiography}[{\includegraphics[width=1in,height=1.25in,clip,keepaspectratio]{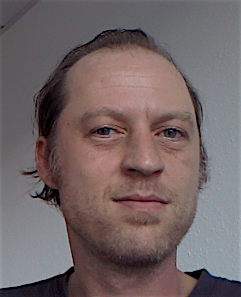}}]{Christian S. Pilz}
Christian S. Pilz (Member, IEEE) was born in Konstanz, Germany, in 1976. He received the German engineering degree (Dipl.Ing.) in media and computer sciences at Giessen, Germany and the M.Sc. in computer sciences at Munich, Germany. He is inventor of more than a dozen patents in the field of speech processing and pattern recognition. He is author of several IEEE CVPR and ICCV papers. He passed his Ph.D. studies in Aachen, Germany.  Currently, he\textquotesingle s working as a research engineer at CanControls GmbH in Aachen, Germany and as a visiting scholar at the Department of Psychiatry, Psychotherapy and Psychosomatics, University
Hospital, Aachen, Germany. 
\end{IEEEbiography}


\begin{IEEEbiography}[{\includegraphics[width=1in,height=1.25in,clip,keepaspectratio]{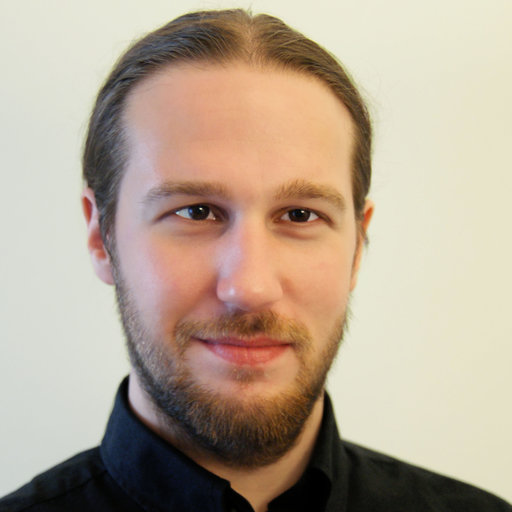}}]{Benjamin Clemens}
Benjamin Clemens, was born in Cologne, Germany, in 1984. He received his M.
Sc. in Psychology from Maastricht University in 2008 and his PhD (Dr. rer.
medic.) from RWTH Aachen University in 2012. He currently works as a Post-
Doctoral researcher and project manager at the Department of Psychiatry,
Psychotherapy   and   Psychosomatics   at   Uniklinik   RWTH   Aachen. He is
additionally  affiliated with the Institute  of Neuroscience and Medicine 
“Brain structure-function relationships” at the Research Centre Jülich. He has
(co)-authored more than 20 research articles in international, peer-reviewed
scientific journals, more than 7 years of teaching experience at the Medical
Faculty of the RWTH Aachen.
\end{IEEEbiography}

\begin{IEEEbiography}[{\includegraphics[width=1in,height=1.25in,clip,keepaspectratio]{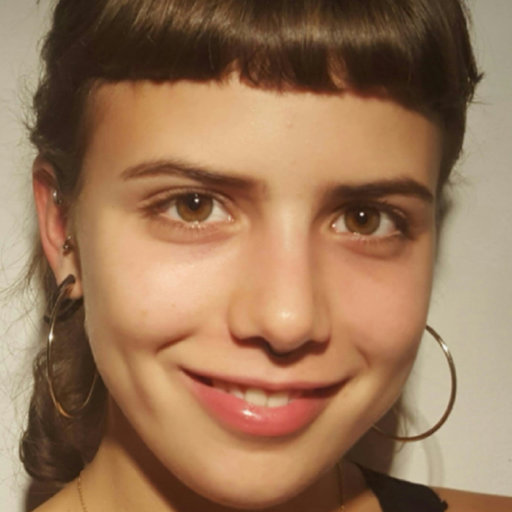}}]{Inka Hiss}
Inka, C. Hiß was born in Breisach Germany, in 1995. She received the B.Sc. in Mental Health and the M. Sc. in Neuropsychology at Maastricht University, Netherlands. Currently, she is doing her Ph.D. at RWTH Aachen University, Germany and is working as a research associate at the Department of Psychiatry, Psychotherapy and Psychosomatic, University Hospital, Aachen, Germany.
\end{IEEEbiography}

\begin{IEEEbiography}[{\includegraphics[width=1in,height=1.25in,clip,keepaspectratio]{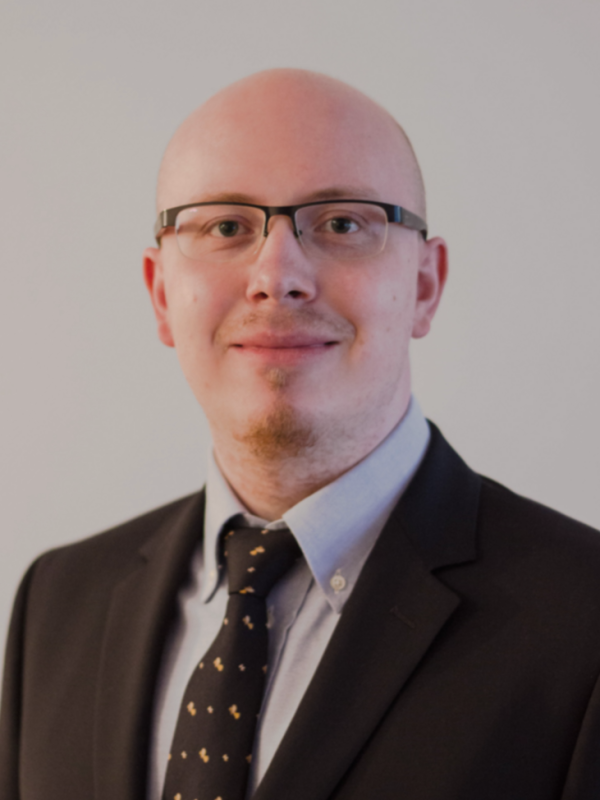}}]{Christoph Weiss}
Christoph Weiss was born in Bonn, Germany, in 1988. He received the Bachelor’s and Master’s degrees in electrical engineering from RWTH Aachen University, Germany, in 2012 and 2016, respectively. He is currently working toward the Ph.D. degree at the Medical Information Technology, RWTH Aachen University, Aachen, Germany. His current research interests include the areas of signal and image processing.
\end{IEEEbiography}

\begin{IEEEbiography}[{\includegraphics[width=1in,height=1.25in,clip,keepaspectratio]{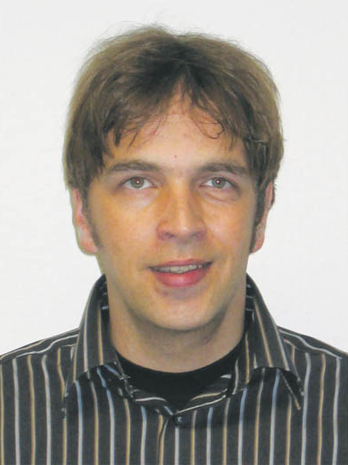}}]{Ulrich Canzler}
 was born in Hannover, Germany, in 1971. He received the German engineering degree (Dipl.-Ing.) in technical computer sciences from the RWTH Aachen, Aachen, Germany and the Dr.-Ing. degree (Ph.D.) in electrical engineering from the RWTH Aachen, Aachen, Germany. He is initiator and author of several public funded medium and larger scale research projects in the field of driver assistant monitoring and human behavior modeling. He is founder and CEO of CanControls GmbH, Aachen, Germany. CanControls GmbH, Aachen, Germany is a Tier 1 supplier in the transport sector shaping the next generation of mobility.
\end{IEEEbiography}

\begin{IEEEbiography}[{\includegraphics[width=1in,height=1.25in,clip,keepaspectratio]{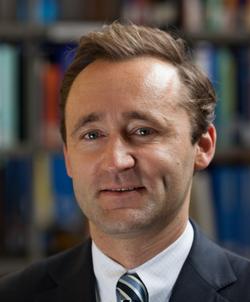}}]{Jarek Krajewski}
Jarek Krajewski received the diploma in 2004 and his doctoral degree for his study on Acoustic Sleepiness Detection in 2008, both in psychology and signal processing from University of Wuppertal and RWTH Aachen in Germany. He is full professor in engineering psychology with the Rhenisch University of Applied Science Cologne, was associate professor in Industrial and Organizational Psychology in Würzburg, and Cologne and vice director of the Center of Interdisciplinary Speech Science at the University of Wuppertal. He is member of the ISCA, Human Factors and Ergonomics Society, German Society of Psychology, and (co-) authored more than 70 publications (h-Index 16, i10-Index 26, 1082 Citations) in peer reviewed books, journals, and conference proceedings in the field of sleepiness detection, and psychophysiological signal processing.
\end{IEEEbiography}


\begin{IEEEbiography}[{\includegraphics[width=1in,height=1.25in,clip,keepaspectratio]{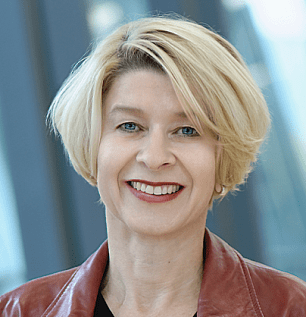}}]{Ute Habel}
Ute Habel was born in Banat, Rumania, in 1969. She received her Diploma in Psychology in 1995 and her PhD (Dr. rer. soc.) in 1998, both from the University of Tübingen. She completed her habilitation at the Faculty of Psychology, University of Vienna, in 2005. Currently, Prof. Habel holds the positions of Leading Psychologist and Head of the Section Neuropsychology at the Department of Psychiatry, Psychotherapy and Psychosomatics at Uniklinik RWTH Aachen, Director of the Institute of Neuroscience and Medicine “Brain structure-function relationships” at the Research Centre Jülich, and Vice Rector for International Affairs of RWTH Aachen University. She has (co)-authored more than 260 publications in international, peer-reviewed scientific journals, her h-Index is 67 and she is the speaker of the International Research Training Group “The Neuroscience of Modulation Aggression and Impulsitivity in Psychopathology“ (DFG, IRTG 2150).
\end{IEEEbiography}

\begin{IEEEbiography}[{\includegraphics[width=1in,height=1.25in,clip,keepaspectratio]{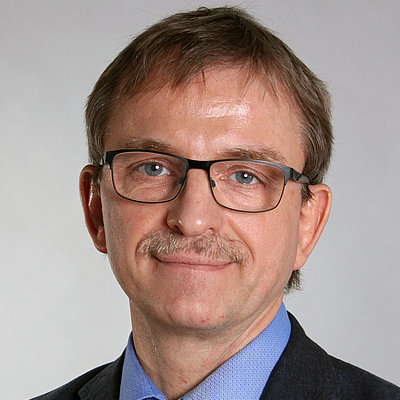}}]{Steffen Leonhardt}
Steffen Leonhardt (Senior Member, IEEE) was born in Frankfurt, Germany, in 1961. He received the M.S. degree in computer engineering from the University at Buffalo, Buffalo, NY, USA, in 1987, the Dr.-Ing. degree (Ph.D.) in electrical engineering from the Technical University of Darmstadt, Darmstadt, Germany, in 1995, the M.D. degree in medicine from J. W. Goethe University, Frankfurt, in 2000, and the Dr. h. c. degree from Czech Technical University in Prague, Prague, Czech Republic, in 2018.
He was appointed as a Full Professor and the Head of the Philips Endowed Chair of Medical Information Technology, RWTH Aachen University, Aachen, Germany, in 2003. He was appointed as a Distinguished Professor at IIT Madras, Chennai, India, in 2018.
Dr. Leonhardt serves as an Associate Editor for the IEEE Journal of Biomedical and Health Informatics and IEEE Transactions on Biomedical Circuits and Systems.
\end{IEEEbiography}




\end{document}